\documentclass[12pt]{article}

\usepackage[margin=1in]{geometry}
\usepackage{amsmath, amssymb, amsthm}
\usepackage{natbib}
\usepackage{hyperref}
\usepackage{graphicx}
\usepackage{booktabs}
\usepackage{microtype}
\usepackage{setspace}
\usepackage{booktabs} % For \toprule, \midrule, \bottomrule
\usepackage{makecell}   % For \makecell
\usepackage{multirow}   % For \multirow
\usepackage{array}
\usepackage{arydshln}   % For \cdashline
\usepackage{enumitem,kantlipsum}

\usepackage{tikz}
\usetikzlibrary{shapes.geometric, arrows.meta, positioning}
\usetikzlibrary{fit, calc}
\usepackage{xcolor}
\usepackage{float}
\usepackage{soul}
\usepackage{color}
\usepackage{helvet}
\usepackage{tcolorbox}
\usepackage{chngcntr}

\soulregister\citep7

\usepackage{natbib}
 \bibpunct[, ]{(}{)}{,}{a}{}{,}%

\theoremstyle{definition}

\title{CLVAE: A Variational Autoencoder for Long-Term Customer Revenue Forecasting}

\author{
    First Author\thanks{Affiliation 1. E-mail: \texttt{email@domain.com}.} \and
    Second Author\thanks{Affiliation 2. E-mail: \texttt{email@domain.com}.}
}

\author{Jeffrey N{\"a}f$^1$,  Riana Valera Mbelson, Markus Meierer$^2$ \\
        $ ^1$Research Institute for Statistics and Information Science, University of Geneva\\
        $ ^2$Institute of Management, University of Geneva}

\date{}

\begin{document}

\maketitle

\begin{abstract}
 Predicting customers’ long-term revenue from sparse and irregular transaction data is central to marketing resource allocation in non-contractual settings, yet existing approaches face a trade-off. Traditional probabilistic customer base models deliver robust long-horizon forecasts by imposing strong structural assumptions, while flexible machine-learning models often require substantial training data and careful tuning. We propose a variational-autoencoder-based model that preserves the process-based likelihood of established attrition–transaction–spend models conditional on customer heterogeneity, but replaces the restrictive parametric mixing distribution with a flexible latent representation learned by encoder-decoder networks. The resulting approach (i) provides a single model for customer attrition, transactions and spending, (ii) remains reliable when contextual covariates are unavailable, and (iii) flexibly incorporates rich covariates and nonlinear effects when they are available. This design balances structural stability with the flexibility needed to capture complex purchase dynamics. Across multiple real-world datasets and prediction horizons, the proposed model improves upon the latest benchmarks. Businesses benefit directly, as a better assessment of customers’ future revenues improves the efficiency of campaign targeting. For research, this work provides guidance on how to embed domain-specific models into the variational autoencoder framework, enabling flexible representation learning while retaining an econometrically meaningful process structure.
\end{abstract}


\bigskip

\section{Introduction}\label{sec:Intro}

\begin{figure}[htbp]
    \centering
    \includegraphics[trim=0.5cm 0cm 0.5cm 6cm, clip, width=\linewidth]{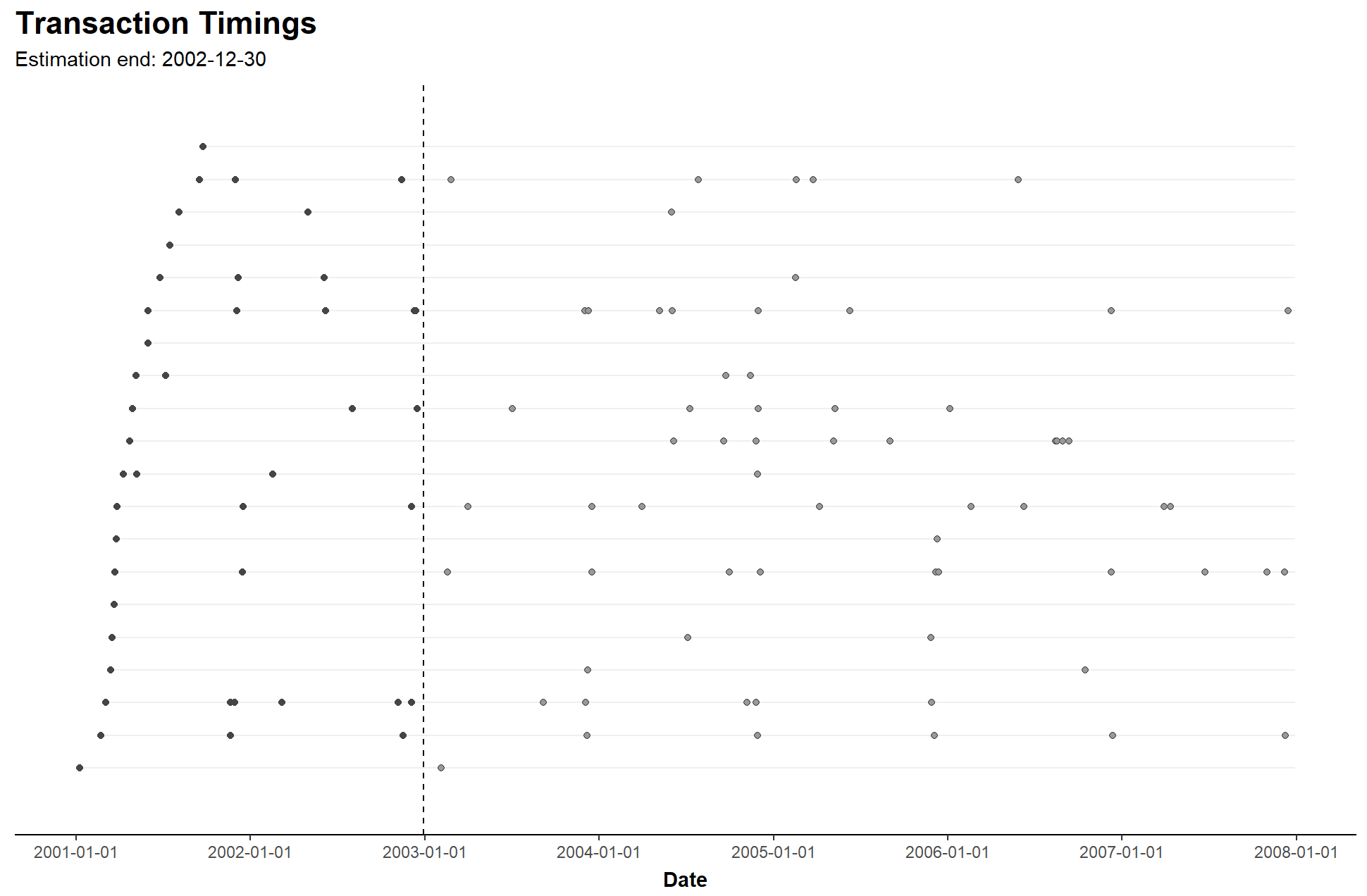}
    \begin{flushleft}
        \footnotesize
        Note: The dashed line indicates the end of a 104-week estimation period. Customers start purchasing at different times during this period. Transactions after the dashed line are part of the prediction period. Not shown is the spend associated with each transaction. 
    \end{flushleft}
    \vspace{0.2cm}    
    \caption{Real-world transaction data for 20 customers (Retailer A, see Section \ref{sec:EmpiricalApplication}).}
    \label{fig:transactiondata}
\end{figure}

% General modeling challenges
Firms in non-contractual settings must routinely infer customers’ long-term future revenue from transaction data that record only purchase timing and monetary value. A central challenge is heterogeneity. Customers differ substantially in their underlying purchase propensities, spending levels, and attrition propensities, yet these differences are only indirectly reflected in sparse transaction data. Because attrition itself is not directly observed, a period without purchases is inherently ambiguous as a customer may be temporarily inactive or may have permanently discontinued purchasing. The structure of the observed transaction records makes learning these differences difficult. In addition, purchase behavior is highly heterogeneous, with up to 50\% of customers purchasing only once while others transact repeatedly. Even among repeat buyers, transaction records are sparse and irregular. Purchases occur at uneven intervals and are separated by long stretches with no transactions (see Figure \ref{fig:transactiondata}). Finally, observation windows vary with customer tenure and are often short relative to the forecasting horizon. As a result, the observed records contain limited information to infer customer-specific propensities and predict long-run revenue.

% Figure with transaction patterns from 20 random customers

% Current best practices and limitations
% --> Cite here in a later stage: CLVTools and EGG paper
A popular approach to address these challenges is to use probabilistic modeling techniques. By disentangling customers' attrition, transaction, and spending processes, these probabilistic models can infer individual customers' future revenue for arbitrary forecasting horizons. Due to their proven robustness and wide accessibility through open-source software \citep{BTYD}, such models are widely used in industry. Common examples are the Pareto/NBD model \citep{Schmittlein1987} for capturing attrition and transactions and the Gamma-Gamma (GG) model \citep{Colombo1999, Fader2005b} for estimating each customer's average spend per transaction. If available, the latter may also be used to model profits \citep{Glady2009}. Probabilistic models that combine all three processes have been proposed by several authors, but their computational complexity and limited improvements in predictive performance have prevented their adoption by practitioners. Although progress has been made to increase the flexibility of these models \citep[e.g.,][]{Abe2009, GladyLemmensCroux2015, PlatzerReutterer2016, BachmannMeiererNaef2021}, these extensions still rely on parametric assumptions about customer heterogeneity that can be too restrictive.

% New possibilities with deep learning 
When considering modern machine learning models, the sparsity and irregularity of the time series for each customer make standard approaches hard to apply. One exception is autoregressive deep learning techniques that may help to address the limited flexibility in traditional models. They can model transaction sequences and have shown promise in predicting customers' transaction patterns \citep{ValendinReuttererPlatzerKalcher2022}. Yet, they often require substantial data to learn structure nonparametrically and represent customer heterogeneity only implicitly, which may be unstable with short and sparse transaction histories. Moreover, they do not directly model spending. This limits their usefulness when the managerial objective is to predict long-term customer revenue. Therefore, the question arises of whether it is possible to retain the structural strengths of traditional probabilistic models while relaxing their restrictive assumptions about latent heterogeneity by leveraging deep learning techniques. Leveraging the "best of both worlds" may facilitate a more robust, scalable and accurate prediction of individual customers' long-term revenue.

% What is a VAE? And how can they help?
% With traditional probabilistic models, such as the Pareto/NBD and GG model in mind, variational autoencoders are a deep learning techniques that might enable this feat. 
To this end, we focus on variational autoencoder techniques that build on an encoder-decoder neural network architecture. Generally speaking, a variational autoencoder (VAE) is a generative latent-variable model trained using variational inference. It provides a flexible way to model high-dimensional data \(X\) via low-dimensional latent variables \(Z\) that capture heterogeneity and structure relevant for prediction, compression and simulation. VAEs are increasingly applied in management, marketing and information systems research \citep[e.g.,][]{SisodiaBurnapKumar2025, YangZhangFan2023, BurnapHauserTimoshenko2023, DewAnsariToubia2022}. In this paper, we show that a VAE presents a natural nonparametric extension of traditional probabilistic customer base models with latent heterogeneity, such as the Pareto/NBD model. Since these models essentially compress the observed recency and frequency data (i.e., $x,t_x,T$) to latent variables ($\lambda,\mu$ in the case of the Pareto/NBD model), the mapping fits naturally into an encoder-decoder perspective. This is the driving idea behind our model development. We consider not only attrition and transaction timing, but also the spending process, inspired by the GG model. Specifically, we keep the likelihood of the combined Pareto/NBD and GG model conditional on the latent heterogeneity variables, but replace the fixed parametric mixing distribution with a latent representation learned by encoder-decoder networks and estimated via amortized variational inference. %This yields controlled flexibility, i.e., the process-based likelihood provides stability, while the learned heterogeneity distribution captures atypical purchasing and spending patterns.

% Contributions
Central to the resulting Customer Lifetime Value Variational Autoencoder -- for short, CLVAE -- is a likelihood-based VAE that preserves the canonical attrition–transaction–spend process likelihood conditional on customer heterogeneity, while learning a flexible nonparametric mixing distribution over customer-level propensities. Our proposed approach makes five contributions relative to prior work: First, it provides a single model to capture all three key processes and their dependencies, which facilitates the direct estimation of customer revenue. Second, it relaxes restrictive parametric heterogeneity assumptions by learning a flexible heterogeneity distribution, enabling latent propensities across attrition, transaction timing, and spending. Third, it uses a fast estimation routine based on amortized inference \citep{Kingma2019VAE}, which supports both moderately sized datasets and large-scale customer bases. Fourth, it is data-parsimonious, requiring only summary statistics of transaction records, yet it can incorporate richer inputs such as full transaction histories and contextual covariates when available. Fifth, it delivers robust and accurate results without extensive dataset-specific tuning. Collectively, these contributions address the main practical trade-off in long-term revenue forecasting by balancing structural stability under sparse transaction histories with flexibility in modeling heterogeneity, while remaining scalable for operational use.

% Results of benchmarking 
We build on multiple real-world datasets to evaluate our approach. The results show that it is scalable, robust, and accurate. In out-of-sample testing across three non-contractual business settings and four prediction horizons, we find that the novel model improves upon the latest benchmarks. To illustrate its managerial relevance and ability to account for additional contextual information, we also show how CLVAE can serve as a practical alternative to the common industry workflow of fitting separate models by acquisition cohort to accommodate observed heterogeneity.

% Remainder of the paper 
The remainder of the paper is structured as follows. Section~\ref{sec:Literature} reviews related research. Section~\ref{sec:Model} introduces the model as well as the estimation and prediction routine. Additionally, it details extensions that broaden applicability across business settings. Section~\ref{sec:EmpiricalApplication} reports on our empirical applications, providing an overview of the datasets, benchmarks, and empirical results. In Section~\ref{sec:Discussion}, we conclude with implications for marketing research and practice. To facilitate replication and further research, we have released a Python implementation of our novel model along with code to reproduce our results in a public repository (\emph{link removed to ensure anonymous review}).

%%%%%%%%%%%%%%%%%%%%%%%%%%%%%%%%%%%%%%%%%%%%%%%%%%%%%%%%%%%%%%%%%%%%%%
%%%%%%%%%%%%%%%%%%%%%%%%%%%%%%%%%%%%%%%%%%%%%%%%%%%%%%%%%%%%%%%%%%%%%%
%%%%%%%%%%%%%%%%%%%%%%%%%%%%%%%%%%%%%%%%%%%%%%%%%%%%%%%%%%%%%%%%%%%%%%

\section{Literature}\label{sec:Literature}

% General motivation for these models 
Customer base analysis is key to many customer-centric initiatives. These initiatives rely on effective customer targeting, which often builds on long-term projections of customer behavior \citep[e.g.,][]{LemmensGupta2020, Simester2006}. While many statistical techniques are applicable for short-term forecasting, data limitations often restrict their direct application for long-term revenue prediction. This is particularly true when transaction records are sparse and the prediction horizon exceeds the available observation window.

% Summary Literature Section
Our work sits at the intersection of (i) probabilistic customer base models for long-term revenue forecasting in non-contractual settings and (ii) modern machine-learning methods for transaction data. The core tension is that long-horizon prediction benefits from strong structural discipline when transaction histories are limited and unevenly observed. At the same time, this discipline typically requires restrictive assumptions about unobserved heterogeneity. We review these streams and highlight the gap that the proposed model addresses.

%%%%%%%%%%%%%%%%%%%%%%%%%%%%%%%%%%%%%%%%%%%%%%%%%%%%%%%%%%%%%%%%%%%%%%

\subsection{Customer base models for long-term forecasting}\label{subsec:CBA_Literature}

% Probabilistic models - Simple
For long-term prediction of customer revenue, probabilistic customer base models are a popular choice. These models generate long-horizon forecasts by distinguishing three processes: customer attrition, transactions, and spending. Their parameter and data parsimony support robustness and broad applicability. Moreover, time-invariant and time-varying covariates can be incorporated, and the resulting coefficients admit standard economic interpretations (e.g., elasticities). Canonical approaches combine a latent attrition model, such as the Pareto/NBD model \citep[e.g.,][]{Schmittlein1987, Fader2005c, Abe2009, Fader2010, BemmaorGlady2012, BraunSchweidelStein2015} with a spending model such as the GG model \citep[e.g.,][]{Colombo1999, Fader2005b, schmittlein1994customer}. These models remain widely used because they provide stable forecasts from limited transaction histories. At the same time, that stability is achieved by committing to specific parametric assumptions about customer heterogeneity and the links between the three processes. By shrinking individual parameters toward an assumed prior distribution, these models remain stable in sparse histories. However, misspecification can turn that shrinkage into bias and harm long-horizon forecasts.

% Probabilistic models - Advanced
%However, in certain settings—such as gift data, where seasonality introduces sharp peaks and extreme observations—the population average can be biased. As a result, inference and forecasting may be distorted, making classic models sensitive to outliers and extreme values.
%Standard implementations rely on fixed parametric specifications, and model misspecification can degrade forecasts, especially when the prediction horizon is long. 
Prior work has therefore introduced joint models that link attrition, transactions, and spending and allow dependence across processes \citep[e.g.,][]{BorleSinghSiddharthJain2008, Schweidel2013, GladyLemmensCroux2015}, as well as extensions that increase flexibility in the underlying structure and with regards to contextual factors \citep[e.g.,][]{PlatzerReutterer2016, BachmannMeiererNaef2021}. While these models provide a richer behavioral process specification, gains in predictive performance have often been modest, whereas computational burden and sensitivity to estimation choices can increase substantially. As a result, stability and scalability may be insufficient for routine, large-scale operational use, which may help explain their limited practitioner adoption.

% ML / Deep Learning techniques
Modern machine-learning methods address some limitations by offering flexible function approximation. These end-to-end-trainable models automatically learn task-relevant features from high-dimensional data, capture complex nonlinearities and interactions, scale to large datasets, and deliver high predictive accuracy \citep[e.g.,][]{Gabel2022}. In case of customer base analysis, research has shown that traditional probabilistic latent attrition models can be replaced by autoregressive sequence models to predict transactional incidence patterns. Long Short-Term Memory (LSTM) models have been a popular choice for this \citep{ValendinReuttererPlatzerKalcher2022}. In sparse settings, this flexibility can overfit without careful regularization. Moreover, these methods often represent heterogeneity only implicitly through high-capacity architectures and focus on modeling transaction incidence, with spending handled in a separate model.

% Summary of gap in literature
Taken together, the literature suggests a persistent gap. Practitioners need models that preserve the long-horizon stability of likelihood-based customer base approaches while relaxing restrictive parametric assumptions, without sacrificing scalability and robustness on sparse and irregular transaction data. Combining deep learning techniques with traditional probabilistic modeling can provide a solution to this challenge. Work in other disciplines has shown the potential of VAE approaches to facilitate such synergies \citep[e.g.,][]{Krishnan2017}.

%%%%%%%%%%%%%%%%%%%%%%%%%%%%%%%%%%%%%%%%%%%%%%%%%%%%%%%%%%%%%%%%%%%%%%

\subsection{Variational autoencoders in management, marketing and information systems}\label{sub:VAE_Marketing}

% Definition VAE
VAEs are likelihood-based latent-variable models estimated with amortized variational inference \citep{Kingma2019VAE}.
Amortized variational inference uses a single recognition model \(q_{\phi}(Z \mid X)\) with parameters \(\phi\) that maps each observation \(X\) to the parameters of an approximate posterior over latent factors \(Z\).
This replaces solving a separate variational problem for every observation and enables fast inference for a new \(X\) via a single forward pass.
The generative model \(p_{\theta}(X \mid Z)\) with parameters \(\theta\) and the distribution \(p(Z)\) define the data likelihood and the prior respectively. The parameters $\theta$, $\phi$ typically parametrize two neural networks, allowing for highly complex, nonlinear relations. Unlike discriminative models that require labeled outputs, e.g., recurrent neural networks for sequence prediction, VAEs learn probabilistic latent representations in an unsupervised or semi-supervised manner, which is valuable for problems where formal labels are lacking.

%On the one hand, this framework preserves the discipline of generative statistics, including likelihoods, priors, posteriors, and posterior-predictive checks.
%On the other hand, it uses deep encoders and decoders to capture nonlinear structure. 

% Literature Overview
 VAEs are widely used across disciplines, and the literature in management, marketing, and information systems is seeing a growing number of VAE-based applications. Existing studies are mainly concentrated in three areas: (i) recommender systems, where the objective is ranking or personalization, (ii) marketing design with high-dimensional inputs such as images and (iii) text applications that learn predictive, managerially interpretable representations. 
In recommender systems, \citet{Liang2018} propose a multinomial VAE for implicit-feedback that uncovers nonlinear user-item structure, improving personalization from sparse click and purchase histories. Similarly, \citet{Bauman2025} learn latent representations for context-aware recommendation and \citet{BoughanmiAnsariLi2025} use message-passing VAEs on interlocked hypergraphs to model categorized consumer collections (e.g., playlists) and learn embeddings for consumers, items, and categories.
In marketing design applications, studies focus on high-dimensional inputs and design exploration. \citet{DewAnsariToubia2022} develop a multimodal VAE aligning logo images with textual brand descriptors to support counterfactual analyses of design choices and brand perceptions. In short-form video marketing, \citet{TianDewIyengar2024} summarize TikTok content into latent video quality representations for causal influencer selection. In new product development, \citet{BurnapHauserTimoshenko2023} combine a probabilistic VAE with adversarial training to learn a controllable latent space of product aesthetics that predicts appeal and generates novel designs. Extending these ideas to decision-centric conjoint, \citet{SisodiaBurnapKumar2025} recover interpretable latent visual factors from images and deploy them for visual conjoint analysis. Broadening beyond recommendation and vision, \citet{YangZhangFan2023} integrate a VAE with task supervision to produce predictive, managerially interpretable topics from text.

Taken together, this literature shows that VAE-based representations can capture complex, nonlinear structure beyond traditional parametric models. A common theme is that this added flexibility can improve predictive performance when hand-crafted features or linear latent-factor models are insufficient. However, prior work has not examined VAEs in the context of customer base analysis. That is, using VAEs to model customer heterogeneity from sparse and irregular transaction histories in order to support long-term revenue prediction in non-contractual settings. More broadly, the paper contributes a modeling template for embedding domain-specific, process-based likelihoods into VAEs, enabling flexible representation learning without discarding econometrically meaningful process structure.

%%%%%%%%%%%%%%%%%%%%%%%%%%%%%%%%%%%%%%%%%%%%%%%%%%%%%%%%%%%%%%%%%%%%%%
%%%%%%%%%%%%%%%%%%%%%%%%%%%%%%%%%%%%%%%%%%%%%%%%%%%%%%%%%%%%%%%%%%%%%%
%%%%%%%%%%%%%%%%%%%%%%%%%%%%%%%%%%%%%%%%%%%%%%%%%%%%%%%%%%%%%%%%%%%%%%

\section{Model}\label{sec:Model}

% Summary Model Section
Following our review of the relevant literature, we now derive our novel VAE-based approach. Thereby, we first detail the structure of the new model and then the respective estimation and prediction routine. While the first three sections detail the basic model structure, we subsequently illustrate how to extend this basic model structure, i.e., by adding covariates and by making more extensive use of customers' transaction history.  Appendices \ref{app:appendix_1} to \ref{app:appendix_4} accompany the discussion in this section.

%%%%%%%%%%%%%%%%%%%%%%%%%%%%%%%%%%%%%%%%%%%%%%%%%%%%%%%%%%%%%%%%%%%%%%

\subsection{Model specification}
\label{sub:model_spec}

% Figure model specification
\begin{figure}[b!]
\begin{center}
\includegraphics[clip, trim=1.2cm 0.4cm 1.2cm 0.4cm, width=16.5cm]{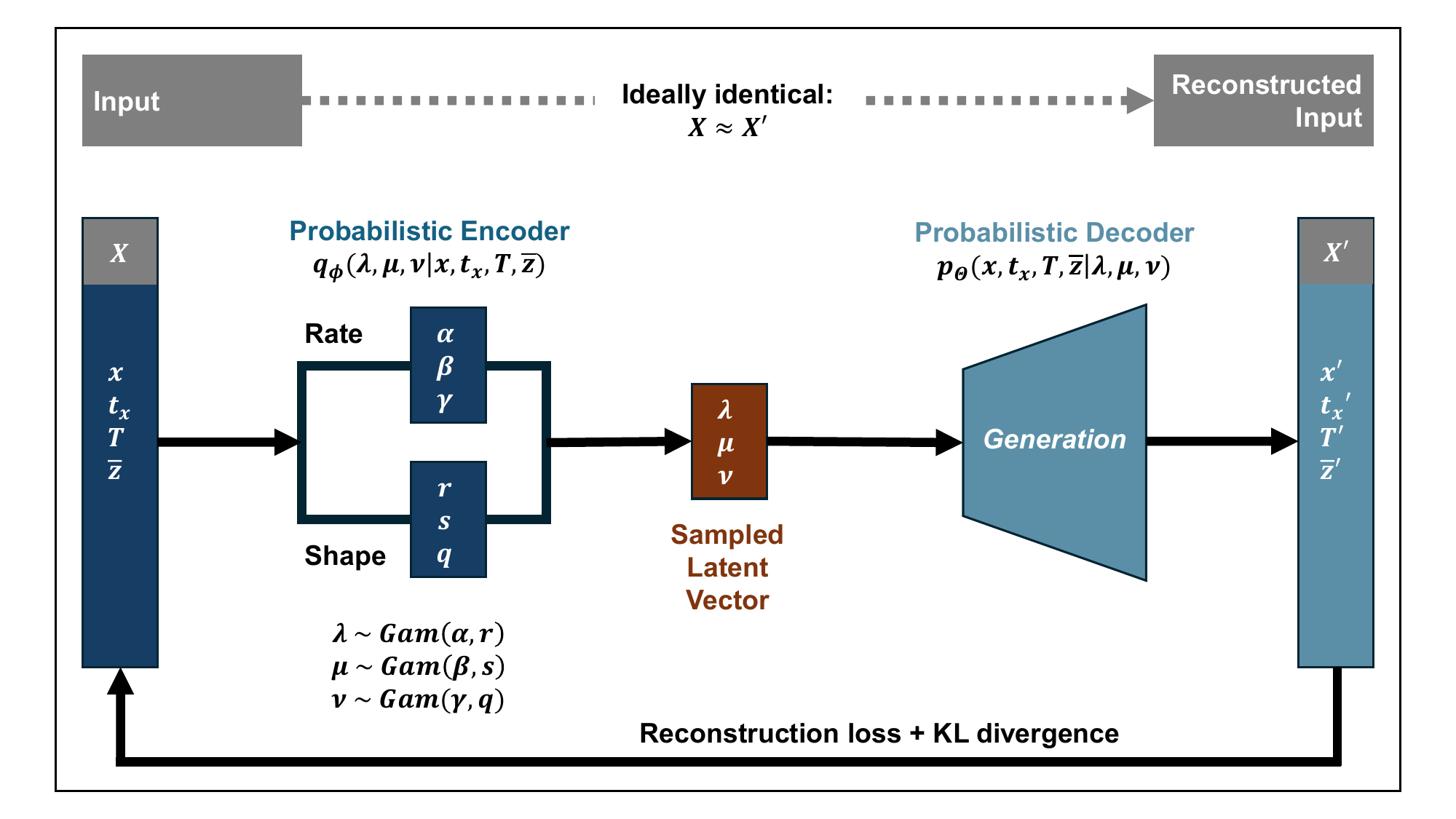}
\caption{Conceptual Visualization of the CLVAE model}
\label{Modelview}
\end{center}
\end{figure}

Building on the definition given in Section \ref{sub:VAE_Marketing}, we note that a VAE is a neural network technique that employs a Bayesian modeling paradigm, in which we model variables $X$ of dimension $D$ using latent variables $Z$ of smaller dimension $d$ \citep{KingmaWelling2014}. In doing so we must specify the prior $p(Z)$, which usually does not depend on parameters, and $p_{\theta}(X \mid Z)$, depending on a neural net parametrized by $\theta$. Although it might also be possible to obtain the posterior $Z \mid X$ in a VAE, this is approximated by another distribution $q_{\phi}(Z \mid X)$. Note that in most applications, $p_{\theta}(X \mid Z)=N(\mu_{\theta}(Z), \Sigma_{\theta} (Z))$, where $\mu_{\theta}, \Sigma_{\theta}$ are functions learned by a neural net with parameters $\theta$. However, our work takes a different route by adapting the VAE framework to the Pareto/NBD and the GG model. In this model, $X=(x,t_x, T, \bar{z})$ are observed for each customer -- i.e., the indicators for customers' recency, frequency, and monetary value (RFM) -- and then modeled as latent variables $Z=(\lambda, \mu, \nu)$. They are sometimes referred to as latent heterogeneity variables. As such, the Pareto/NBD and the GG model can naturally be embedded into a VAE-based framework.

Figure \ref{Modelview} illustrates our modeling framework, whereby $p_{\theta}(x,t_x, T, \bar{z} \mid \lambda, \mu, \nu)$ is specified as,
\begin{align}
\label{eq:likelihood}
% &p_{\theta}(X \mid Z) \nonumber \\
% &=p_{\theta}(x,t_x, T, \bar{z} \mid \lambda, \mu, \nu) \nonumber\\
% &= \Big(\frac{\Lambda_{\theta}(\lambda,\mu, \nu)^x M_{\theta}(\lambda,\mu, \nu)}{\Lambda_{\theta}(\lambda,\mu, \nu) + M_{\theta}(\lambda,\mu, \nu)} e^{-(\Lambda_{\theta}(\lambda,\mu, \nu) + M_{\theta}(\lambda,\mu, \nu))t_x} \nonumber \\
% &+ \frac{\Lambda_{\theta}(\lambda,\mu, \nu)^{x+1}}{\Lambda_{\theta}(\lambda,\mu, \nu) + M_{\theta}(\lambda,\mu, \nu)} e^{-(\Lambda_{\theta}(\lambda,\mu, \nu) + M_{\theta}(\lambda,\mu, \nu))T}\Big)\nonumber\\
% &\frac{(N_{\theta}(\lambda,\mu, \nu) x)^{px}}{\Gamma(px)} \bar{z}^{px-1} e^{-N_{\theta}(\lambda,\mu, \nu) x \bar{z}}.
% %
&p_{\theta}(X \mid Z) \nonumber \\
&=p_{\theta}(x,t_x, T, \bar{z} \mid \lambda, \mu, \nu) \nonumber\\
&= \Big(\frac{\Lambda_{\theta}(\lambda,\mu, \nu)^x M_{\theta}(\lambda,\mu, \nu)}{\Lambda_{\theta}(\lambda,\mu, \nu) + M_{\theta}(\lambda,\mu, \nu)} \nonumber \\
&\ \ \ \ e^{-(\Lambda_{\theta}(\lambda,\mu, \nu) + M_{\theta}(\lambda,\mu, \nu))t_x} \nonumber \\
&\ \ \ \ + \frac{\Lambda_{\theta}(\lambda,\mu, \nu)^{x+1}}{\Lambda_{\theta}(\lambda,\mu, \nu) + M_{\theta}(\lambda,\mu, \nu)} \nonumber\\
&\ \ \ \ e^{-(\Lambda_{\theta}(\lambda,\mu, \nu) + M_{\theta}(\lambda,\mu, \nu))T}\Big)\nonumber\\
&\ \ \ \ \frac{(N_{\theta}(\lambda,\mu, \nu) x)^{px}}{\Gamma(px)} \bar{z}^{px-1} e^{-N_{\theta}(\lambda,\mu, \nu) x \bar{z}}.
\end{align}

Note that if $\Lambda_{\theta}(\lambda,\mu, \nu)=\lambda$, $M_{\theta}(\lambda,\mu, \nu)=\mu$ and $N_{\theta}(\lambda,\mu, \nu)=\nu$, $p_{\theta}(x,t_x, T, \bar{z} \mid \lambda, \mu, \nu)$ is the standard likelihood of the Pareto/NBD and GG model conditioned on the latent variables: That is, equation~\eqref{eq:likelihood} combines the likelihood of the Pareto/NBD model \cite[Equation (14)]{Fader2005a} with the density of the GG model \citep{Fader2013a}. However, now complex interactions between the latent variables are possible, leading to potentially arbitrary dependence, as well as nonstandard distributions for $(\Lambda_{\theta}(\lambda,\mu, \nu), M_{\theta}(\lambda,\mu, \nu), N_{\theta}(\lambda,\mu, \nu))$, even if $\lambda, \mu, \nu$ are sampled from independent Gamma distributions, as in the original Pareto/NBD and GG model.

Following the VAE framework, we then approximate the posterior by using the function 
\begin{align}\label{DecoderLik}
    % &q_{\phi}(Z \mid X) \nonumber \\
    % &=q_{\phi}(\lambda, \mu, \nu \mid x,t_x, T, \bar{z}) \nonumber\\
    % &=\frac{\alpha_{\phi}(X)^{r_{\phi}(X)} \lambda^{r_{\phi}(X)-1} \exp[-\lambda \alpha_{\phi}(X)]}{\Gamma(r_{\phi}(X))}\frac{\beta_{\phi}(X)^{s_{\phi}(X)} \mu^{s_{\phi}(X)-1} \exp[-\mu \beta_{\phi}(X)]}{\Gamma(s_{\phi}(X))}\frac{\gamma_{\phi}(X)^{q_{\phi}(X)} \nu_0^{q_{\phi}(X)-1} \exp[-\nu_0 \gamma_{\phi}(X)]}{\Gamma(q_{\phi}(X))},
    %%%
        &q_{\phi}(Z \mid X) \nonumber \\
    &=q_{\phi}(\lambda, \mu, \nu \mid x,t_x, T, \bar{z}) \nonumber\\
    &=\frac{\alpha_{\phi}(X)^{r_{\phi}(X)} \lambda^{r_{\phi}(X)-1} \exp[-\lambda \alpha_{\phi}(X)]}{\Gamma(r_{\phi}(X))} \nonumber \\
    &\ \ \ \ \frac{\beta_{\phi}(X)^{s_{\phi}(X)} \mu^{s_{\phi}(X)-1} \exp[-\mu \beta_{\phi}(X)]}{\Gamma(s_{\phi}(X))} \nonumber \\
    &\ \ \ \ \frac{\gamma_{\phi}(X)^{q_{\phi}(X)} \nu_0^{q_{\phi}(X)-1} \exp[-\nu_0 \gamma_{\phi}(X)]}{\Gamma(q_{\phi}(X))},
\end{align}
which is simply the product of three independent Gamma distributions, again motivated by the Pareto/NBD and GG model, in which the latent variables are modeled with independent Gamma distributions. Finally, again following the traditional framework of the Pareto/NBD and GG model, we assume the \emph{prior} $p(Z)=p(\lambda, \mu, \nu)$ to be the product of independent Gamma distributions:
\begin{align}\label{prior}
% p(\lambda, \mu, \nu) =
% \frac{\alpha_p^{r_p} \, \lambda^{\,r_p-1} \, e^{-\alpha_p \lambda}}{\Gamma(r_p)}
% \;\;
% \frac{\beta_p^{s_p} \, \mu^{\,s_p-1} \, e^{-\beta_p \mu}}{\Gamma(s_p)}
% \;\;
% \frac{\gamma_p^{q_p} \, \nu^{\,q_p-1} \, e^{-\gamma_p \nu}}{\Gamma(q_p)}.
%%%
&p(\lambda, \mu, \nu) =
\frac{\alpha_p^{r_p} \, \lambda^{\,r_p-1} \, e^{-\alpha_p \lambda}}{\Gamma(r_p)}
\frac{\beta_p^{s_p} \, \mu^{\,s_p-1} \, e^{-\beta_p \mu}}{\Gamma(s_p)} \nonumber \\
&\ \ \ \ \ \ \ \ \ \ \ \ \ \ \ \ \ \ \ \ \frac{\gamma_p^{q_p} \, \nu^{\,q_p-1} \, e^{-\gamma_p \nu}}{\Gamma(q_p)},
\end{align}

where the prior parameters $\alpha_p, r_p, \beta_p, s_p, \gamma_p, q_p$ are fixed, e.g., using the optimized values from a fitted Pareto/NBD and GG model (see also Online Appendix~2).

We note that choosing both the prior $p(\lambda, \mu, \nu)$, as well as the variational approximation to the posterior $q_{\phi}(\lambda, \mu, \nu \mid x,t_x, T, \bar{z})$ as the product of independent Gamma distributions, not only aligns well with the framework of the Pareto/NBD as well as the GG model, but also allows for efficient estimation and prediction, as outlined in the next two sections. However, as mentioned above, the combination of the Pareto/NBD and GG model likelihood in \eqref{eq:likelihood} is then based on $(\Lambda_{\theta}(\lambda,\mu, \nu), M_{\theta}(\lambda,\mu, \nu), N_{\theta}(\lambda,\mu, \nu))$, which share complex dependencies and nonstandard distributions, leading to a true joint model of transaction, attrition and spending. Finally, following the Pareto/NBD and GG model, we only consider the RFM indicators in the proposed model, but we note that various extensions to include further information are possible. Respective model extensions are discussed in Section \ref{sec:Extension1} and \ref{sec:Extension2}.

%%%%%%%%%%%%%%%%%%%%%%%%%%%%%%%%%%%%%%%%%%%%%%%%%%%%%%%%%%%%%%%%%%%%%%

\subsection{Estimation}

Having established the model specification, we now turn to the estimation procedure. Central to our inference strategy is the maximization of the Evidence Lower Bound (ELBO), which provides a tractable variational approximation to the intractable posterior over both the latent variables and the learned hyperparameters. We now detail the ELBO formulation, its decomposition, and the stochastic optimization routine used to train the parameters $(\phi, \theta)$. See also Appendix~\ref{app:appendix_1}.

Following the VAE framework \citep{KingmaWelling2014}, the ELBO can be derived as: 
\begin{align}
\label{eq: elbo_pnbd_vae}
&\mathcal{L}_{\theta, \phi}(X) = \mathbb{E}_{q_\phi(\lambda, \mu, \nu \mid X)} \big[\log p_\theta(X \mid \lambda, \mu, \nu) \nonumber \\
&\ \ \ \ \ \ \ \ \ \ \ \ \ \ \ \ \ - \mathrm{KL}\bigl(q_\phi(\lambda, \mu, \nu \mid X) \,\|\, p(\lambda, \mu, \nu)\bigr)\big], 
\end{align}
where $\mathrm{KL}\bigl(q_\phi(\lambda, \mu, \nu \mid X) \,\|\, p(\lambda, \mu, \nu)\bigr)$ is the Kullback-Leibler (KL) divergence between the variational posterior and the prior for the latent variables $(\lambda, \mu, \nu)$. Hence, the ELBO naturally decomposes into a trade-off between two terms: (a) a likelihood term, which quantifies how well the model, via the variational distribution, explains the observed data under the probabilistic decoder, and (b) a regularization term, which penalizes divergence from the prior distribution over the latent variables. Consequently, optimizing the ELBO enables scalable unsupervised learning of the approximate posterior distribution over the latent variables.  

The first term in the ELBO, $p_\theta(X \mid \lambda, \mu, \nu)$, is given in equation \eqref{eq:likelihood}. Moreover, using equations \eqref{DecoderLik} and \eqref{prior}, the variational posterior and prior factorize over the components as
\begin{align}
&\mathrm{KL}\bigl(q_\phi(\lambda, \mu, \nu \mid X) \,\|\, p(\lambda, \mu, \nu)\bigr) \nonumber
\\
&= \mathrm{KL}\bigl(q_\phi(\lambda \mid X) \,\|\, p(\lambda)\bigr) \nonumber \\
&\ \ \ \ + \mathrm{KL}\bigl(q_\phi(\mu \mid X) \,\|\, p(\mu)\bigr)\nonumber\\ 
&\ \ \ \ + \mathrm{KL}\bigl(q_\phi(\nu \mid X) \,\|\, p(\nu)\bigr).
\end{align}

Each component KL divergence between Gamma distributions with variational parameters \((\alpha_q, \beta_q)\) and prior parameters \((\alpha_p, \beta_p)\) admits the explicit form according to~\citet{Penny2001},
\begin{align}
% &\mathrm{KL}\bigl(\mathrm{Gamma}(\alpha_q, \beta_q) \,\|\, \mathrm{Gamma}(\alpha_p, \beta_p)\bigr) \nonumber \\ 
% &= (\alpha_q - \alpha_p)\, \psi(\alpha_q) - \log \frac{\Gamma(\alpha_q)}{\Gamma(\alpha_p)} + \alpha_p (\log \beta_q - \log \beta_p) + \alpha_q \frac{\beta_p- \beta_q}{\beta_q},
%%%%
&\mathrm{KL}\bigl(\mathrm{Gamma}(\alpha_q, \beta_q) \,\|\, \mathrm{Gamma}(\alpha_p, \beta_p)\bigr) \nonumber \\ 
&= (\alpha_q - \alpha_p)\, \psi(\alpha_q) - \log \frac{\Gamma(\alpha_q)}{\Gamma(\alpha_p)}  \nonumber \\
&\ \ \ \ + \alpha_p (\log \beta_q - \log \beta_p) + \alpha_q \frac{\beta_p- \beta_q}{\beta_q},
\end{align}

where \(\psi(\cdot)\) denotes the Digamma function (see e.g., \cite{abramowitz1972psi}). This allows to efficiently approximate \eqref{eq: elbo_pnbd_vae} using values sampled from \eqref{DecoderLik} and, in turn, to approximately minimize \eqref{eq: elbo_pnbd_vae} using standard stochastic gradient descent methods. We refer to Appendix \ref{app:appendix_2} for further details.

%%%%%%%%%%%%%%%%%%%%%%%%%%%%%%%%%%%%%%%%%%%%%%%%%%%%%%%%%%%%%%%%%%%%%%

\subsection{Prediction}

After fitting the proposed model, we generate predictive distributions for customer-level outcomes via fully vectorized Monte Carlo simulation. For details, see Appendix \ref{app:appendix_3}.

For a customer $i$ with transaction history $x_i, T_i, t_{x_i}$, we seek to predict total spending in period $[T_i, T_i+t]$. Let $\Lambda_{\theta}(\lambda_i,\mu_i, \nu_i)$ and $M_{\theta}(\lambda_i,\mu_i, \nu_i)$ denote the latent purchase and dropout rates, respectively. Let moreover $\Omega_i$ define the latent lifetime of a customer. 

Following the proposed model, note that $\Omega_i \sim \text{Exp}(M_{\theta}(\lambda_i,\mu_i, \nu_i))$ and by the memorylessness property of the exponential 
\begin{align}\label{Omega_draw}
    \Omega_i - T_i \mid \Omega_i > T_i \sim \text{Exp}(M_{\theta}(\lambda_i,\mu_i, \nu_i))
\end{align}

The probability that an individual customer remains active at time $T_i$ was originally derived for the standard Pareto/NBD model by \citet{Schmittlein1987}. It is computed as
\begin{align}
\label{eq:survival_prob}
% &P( \Omega >  T_i \mid X_i, Z_i) \nonumber\\
% &=P( \Omega >  T_i \mid x_i, t_{x_i}, T_i, \Lambda_{\theta}(\lambda_i,\mu_i, \nu_i),M_{\theta}(\lambda_i,\mu_i, \nu_i) )
% \nonumber \\
% &=\Big(1 +\frac{M_{\theta}(\lambda_i,\mu_i, \nu_i)}{\Lambda_{\theta}(\lambda_i,\mu_i, \nu_i) + M_\theta(\lambda_i,\mu_i, \nu_i)} \left( \exp\big[(\Lambda_{\theta}(\lambda_i,\mu_i, \nu_i) + M_{\theta}(\lambda_i,\mu_i, \nu_i))(T_i - t_{x_i})\big] - 1 \right) \Big)^{-1},
%%%%
&P( \Omega >  T_i \mid X_i, Z_i) \nonumber\\
%&=P( \Omega >  T_i \mid x_i, t_{x_i}, T_i, \Lambda_{\theta}(\lambda_i,\mu_i, \nu_i),M_{\theta}(\lambda_i,\mu_i, \nu_i) )
%\nonumber \\
&=\Big(1 +\frac{M_{\theta}(\lambda_i,\mu_i, \nu_i)}{\Lambda_{\theta}(\lambda_i,\mu_i, \nu_i) + M_\theta(\lambda_i,\mu_i, \nu_i)}\nonumber \\
&\ \ \ \ \left( e^{(\Lambda_{\theta}(\lambda_i,\mu_i, \nu_i) + M_{\theta}(\lambda_i,\mu_i, \nu_i))(T_i - t_{x_i})} - 1 \right) \Big)^{-1}.
\end{align}
Note that in our model, this expression depends on the latent parameter of spending, $\nu_i$ as well.

Thus, conditioned on the event that $\Omega_i >  T_i $, dropout times $\Omega_i$ can be sampled using an exponential distribution with rate $M_{\theta}(\lambda_i,\mu_i, \nu_i)$, as in \eqref{Omega_draw}. For each customer deemed alive, purchases are simulated sequentially with inter-purchase times drawn from an exponential distribution with rate $\Lambda_{\theta}(\lambda_i,\mu_i, \nu_i)$,
\begin{equation}
t_{x+j-1,x+j} \sim \text{Exp}(\Lambda_{\theta}(\lambda_i,\mu_i, \nu_i)).
\end{equation}
as long as 
\begin{equation}
    T_i + \sum_{j=1}^{J}t_{x+j-1,x+j} \leq \min(T_i+t, \Omega_i)
\end{equation}
where $t$ is the time horizon. The total number of simulated purchases for customer $i$ in simulation $l$ is denoted by $N_i^{(l)}$, and the point estimate of expected purchase frequency is given by
\begin{equation}
\hat{N}_i = \frac{1}{L} \sum_{l=1}^{L} N_i^{(l)},
\end{equation}
where $L$ is the number of Monte Carlo samples.

Conditional on purchase counts, the total spending amount simulated from a Gamma distribution parameterized by the auxiliary latent variables:
% \begin{equation}
% S_i^{(l)} \sim \text{Gam}(\text{shape} = p\cdot N_i^{(s)}, \; \text{rate} = q\cdot N_i^{(s)}).
% \end{equation}
\begin{equation}
%S_i^{(l)} \sim \text{Gam}(\text{shape} = p\cdot N_i^{(s)}, \; \text{rate} = q).
S_i^{(l)} \sim \text{Gam}( p\cdot N_i^{(l)},  q).
\end{equation}
The expected total spending per customer is obtained as $\hat{S}_i = \frac{1}{L} \sum_{l=1}^{L} S_i^{(l)}$. We note that this simulation also makes it directly possible to generate the full distribution of predicted period spending.

%%%%%%%%%%%%%%%%%%%%%%%%%%%%%%%%%%%%%%%%%%%%%%%%%%%%%%%%%%%%%%%%%%%%%%

\subsection{Model extensions: Adding covariates to account for contextual factors}\label{sec:Extension1}

Adding covariates within the VAE framework is not only straightforward, but also goes beyond what covariate implementations of traditional probabilistic models for customer base analysis can do \citep[e.g.,][]{Abe2009}.

Often, companies have access to relevant contextual information that can be represented by vectors of covariates $Y_i^P, Y_i^A, Y_i^S$ for each customer $i$. Examples include customers' age, gender, or acquisition cohort membership. The latter describes an indicator variable that details when customers started purchasing from a business as these customers have likely been acquired through the same campaign they are often assumed to be similiar in their behavior.  
So far, such covariates were incorporated into the Pareto/NBD and GG model using a proportional hazard approach, whereby individual latent parameters were defined through $\lambda_i=\exp(\gamma_{purch}^{\top} Y_i^P) \lambda$, $\mu_i=\exp(\gamma_{att}^{\top} Y_i^A) \mu$, $\nu_i=\exp(\gamma_{spend}^{\top}Y_i^S) \nu$, with $\lambda \sim \text{Gam}(\alpha, r)$, $\mu \sim \text{Gam}(\beta, s)$ and $\nu \sim \text{Gam}(\gamma, q)$ as before. This effectively rescaled the latent variables for each customer, which is equivalent to drawing the latent variables from rescaled Gamma distributions: $\lambda_i \sim \text{Gam}(\alpha_i, r)$, $\mu_i \sim \text{Gam}(\beta_i, s)$ and $\nu_i \sim \text{Gam}(\gamma_i, q)$, where
\begin{align*}
    &\alpha_i=\exp(-\gamma_{purch}^{\top} Y_i^P) \alpha, \beta_i=\exp(-\gamma_{att}^{\top} Y_i^{A}) \beta, \\
    &\gamma_i=\exp(-\gamma_{spend}^{\top} Y_i^A) \gamma
\end{align*}
Although this approach is powerful, it is also relatively simple and only allows the scale parameters of each Gamma distribution to be affected by the covariates. This might be limiting in many instances. To illustrate this point, we briefly elaborate on the current practice to account for customers' assignment to an acquisition cohort, which is known to be an important contextual factor in customer base analysis. 
If a business conducts a customer base analysis, differences between customer acquisition cohorts are usually accounted for. Current modeling practices mainly offer two ways to do so: (I) using separate Pareto/NBD and GG models for each cohort, or (II)  defining covariates that indicate a customer's acquisition cohort and fit a single Pareto/NBD and GG model with these indicator variables for all cohorts. While (II) is much more sample efficient, the covariates can only influence $\alpha$, $\beta$ and $\gamma$ and do not lead to individual shape parameters. Thus, approach (II) is often less effective than (I). This is also related to the fact that the traditional framework relying on the Pareto/NBD and GG model naturally has limited flexibility, as does any parametric approach. Additionally, if customers from different acquisition cohorts are widely similar, choosing (II) might actually hurt the predictive performance of the model as adding irrelevant covariates often increases the estimation error in the traditional probabilistic modeling framework (see Online Appendix~2).   

Our approach delivers a natural way to extend this framework by letting the covariates enter the probabilistic encoder. That is, we now define for a random customer with covariates $\mathbf{Y}=\left(Y^P, Y^A, Y^S \right)$ the posterior $q_{\phi}(Z \mid X, \mathbf{Y})=q_{\phi}( \lambda, \mu, \nu \mid x,t_x,T, \mathbf{Y})$ using
\begin{align*}
% (r(X, \mathbf{Y}), \alpha(X, \mathbf{Y}), s(X, \mathbf{Y}), \beta(X, \mathbf{Y}), q(X, \mathbf{Y}), \gamma(X, \mathbf{Y})) = f_\phi(X, \mathbf{Y}),
%%%
\begin{pmatrix}
    r(X, \mathbf{Y})\\
    \alpha(X, \mathbf{Y})\\
    s(X, \mathbf{Y})\\
    \beta(X, \mathbf{Y})\\
    q(X, \mathbf{Y})\\
    \gamma(X, \mathbf{Y}))
\end{pmatrix}= f_\phi(X, \mathbf{Y})
\end{align*}
% \begin{align*}
% &p_{\theta}(X \mid Z, \mathbf{Y})\\
% &=p_{\theta}(x,t_x, T, \bar{z} \mid \lambda, \mu, \nu, \mathbf{Y})\\
% &= \Big(\frac{\Lambda_{\theta}(\lambda,\mu, \nu, \mathbf{Y})^x M_{\theta}(\lambda,\mu, \nu,\mathbf{Y})}{\Lambda_{\theta}(\lambda,\mu, \nu, \mathbf{Y}) + M_{\theta}(\lambda,\mu, \nu,\mathbf{Y})} e^{-(\Lambda_{\theta}(\lambda,\mu, \nu,\mathbf{Y}) + M_{\theta}(\lambda,\mu, \nu,\mathbf{Y}))t_x} \\
% &+ \frac{\Lambda_{\theta}(\lambda,\mu, \nu,\mathbf{Y})^{x+1}}{\Lambda_{\theta}(\lambda,\mu, \nu,Y) + M_{\theta}(\lambda,\mu, \nu,\mathbf{Y})} e^{-(\Lambda_{\theta}(\lambda,\mu, \nu,Y) + M_{\theta}(\lambda,\mu, \nu,\mathbf{Y}))T}\Big)\frac{(N_{\theta}(\lambda,\mu, \nu,\mathbf{Y}) x)^{px}}{\Gamma(px)} \bar{z}^{px-1} e^{-N_{\theta}(\lambda,\mu, \nu,\mathbf{Y}) x \bar{z}}.
% \end{align*}

To see the increased flexibility of this approach, consider the generation of the latent parameter $\lambda$: In both cases, we draw $\lambda \sim \text{Gam}(\alpha, r)$. In the case of the standard Pareto/NBD and GG model, we then use the transformation $\exp(\gamma_{purch}^{\top} Y^{P}) \lambda$, changing only the scaling, i.e.,
\[
\exp(\gamma_{purch}^{\top} Y^{P}) \lambda \sim \text{Gam}(\exp(-\gamma_{purch}^{\top} Y^{P}) \alpha, r).
\]
However, in the proposed CLVAE, $\mathbf{Y}$ enters into the Gamma distribution trough the highly flexible encoder, which even if we disregard the effect of $\mu, \nu$, may lead to a complete change in the distribution. As such, the covariates can have a much larger effect on the model, not just changing the scaling but also the shape of the latent distributions. We will leverage this new framework in Section \ref{sec:EmpiricalApplication} to provide a more effective version of approach (II). 

Finally, we note that adding covariates generally increases the computational complexity for any approach. However, the VAE framework can be run on highly optimized computational hardware, which enables fast estimation and prediction - even for models with many covariates (see also Online Appendix~2).

\subsection{Model extensions: Leveraging additional information in customers' transaction data}\label{sec:Extension2}

Our base model follows the guidance of the Pareto/NBD and GG model and only uses the recency, frequency, and monetary indicators $X= (x, t_x, T, \bar{z})$ . In other words, the transaction history of each customer is represented by only four variables. In addition to being data parsimonious, this also has the advantage of making the model more stable over a range of settings. However, in Appendix~\ref{app:appendix_4}, we present an extension in which a bivariate Long Short-Term Memory (LSTM) model \citep{hochreiter1997long} is used to instead encode the full transaction and spending history of customers. We also note that a Transformer-based architecture \citep{Vaswani2017} can be used instead. These model variations have the potential to leverage substantially richer information, but they also require considerably more parameter tuning and validation to achieve strong performance. 

Whether this additional investment is worthwhile depends on the business setting and, in particular, on the extent to which the underlying customer behavior exhibits meaningful temporal structure or additional irregularity that is not accounted for by the RFM data. In the latter case, LSTM- or even Transformer-based architectures can be effective in exploiting such patterns. However, when the data are closer to a memoryless process (e.g., approximating Poisson-type dynamics or i.i.d. purchase events), interpurchase times are largely independent over time. In this setting, an LSTM may primarily fit noise, increasing model variance and the risk of overfitting without improving out-of-sample generalization. Taken together, these considerations motivate offering both model variations, so the practitioner can match the model complexity to the richness of the data. Regardless of which variant is used, the core idea of the proposed modeling approach remains the same, with the architecture merely adapting to the complexity of the data.

In sum, our novel approach provides an innovative way to generalize the traditional framework given by the Pareto/NBD and GG model. It permits both the latent variable distributions and the conditional likelihood to be adaptively modulated by the data through a deep neural network. As a result, the proposed model is capable of learning complex, data-dependent relationships between covariates and latent behavioral characteristics. Thereby, it achieves a degree of modeling flexibility surpassing that afforded by traditional parametric methodologies.

\section{Empirical application}\label{sec:EmpiricalApplication}

% Summary Empirical Application Section
The empirical assessment of the proposed neural network model is based on the analysis of multiple real-world datasets. A description of these datasets follows in the next section. Then, we apply the proposed model to these datasets and evaluate its accuracy for customer revenue projections. Throughout, we evaluate predictions of customer-level cumulative realized revenue during each holdout period, computed directly from the holdout transactions. Thereby, we also provide a comparison with key benchmark models. All models are estimated using only transactions within the estimation window. The holdout window is used exclusively for out-of-sample evaluation, and no features or preprocessing steps use information from the holdout period. We report results for up to four holdout horizons, measured from the end of the estimation window, to reflect use cases with different forecasting requirements (52, 104, 156, 208 weeks). See Table \ref{table:descriptives} for details. 

% Benchmark models
The proposed model is applied in two variations, with and without covariates. As both cases are fairly common in industry applications, this allows assessing the model's ability to deliver reliable results in either case. The model implementation was executed using the PyTorch framework \citep{Paszke2019}. For a detailed description of the model training, we refer to Appendix~\ref{app:appendix_2}. Notably, the CLVAE models are trained with the same hyperparameters across datasets. This avoids confounding performance differences due to dataset-specific hyperparameter tuning. As a consequence, we interpret the results as conservative performance estimates that emphasize comparability and robustness over dataset-specific optimization. As mentioned above, Appendix~\ref{app:appendix_4} also presents a version of the model using the full transaction history, instead of the more parsimonious RFM data.

We compare the proposed model to three alternatives. We deliberately select benchmarks that are widely used in practice (PNBD + GG), represent a competitive deep-learning baseline (LSTM + GG), and reflect state-of-the-art joint likelihood modeling (PNBDGGII). First, the traditional approach relying on the Pareto/NBD model with and without covariates \citep{Schmittlein1987, Abe2009, BachmannMeiererNaef2021} combined with the Gamma-Gamma model \citep{Colombo1999, Fader2005b}. Due to its wide accessibility through open-source software \citep[e.g.,][]{BTYD}, this is likely the most used modeling approach in industry. Second, the LSTM model by \citet{ValendinReuttererPlatzerKalcher2022} combined with the Gamma-Gamma model. The LSTM model not only has the ability to include covariates but the implementation provided by the original authors also derives and includes seasonal regularities and thereby constitutes a particularly strong benchmark. Further, we use the hyperparameters recommended by the authors. Third, we implemented an improved version of the joint model proposed by \citet{GladyLemmensCroux2015} that builds on the foundations of the Pareto/NBD and GG model but also captures inter-customer and intra-customer correlations. This model also allows to include covariates in all three processes. For details, see Online Appendix~3.

% Pointer to Replication Package
% Replication materials to reproduce the reported results for the proposed approach are available in a public code repository (\emph{link removed to ensure anonymous review}).

%%%%%%%%%%%%%%%%%%%%%%%%%%%%%%%%%%%%%%%%%%%%%%%%%%%%%%%%%%%%%%%%%%%%%%

\subsection{Data}

This study features three real-world datasets that have been selected to cover a broad spectrum of commonly observed customer purchase behavior. This also applies to data availability in terms of covariates. The datasets originate from different industries and include purchase records from three retailers: Retailer A \citep{DEMFGift}, Retailer~B \citep{Ni2012}, and Retailer~C \citep{BookStore}. This ensures that the benchmarking sufficiently accounts for the robustness of each model across a variety of scenarios. 

In addition to this reasoning, a key criterion for selecting these datasets was also their accessibility. All datasets are publicly available, which facilitates replication of the reported results. Because covariate information in public datasets is often limited, proprietary datasets may yield better predictive performance by providing richer contextual variables. Balancing the trade-off between using public and proprietary datasets, our benchmarking setup highlights the practical importance of methods that remain reliable when only transaction histories are available, while still providing a pathway to leverage richer covariates when firms have them. 

\begin{table}[h]
\caption{Key Descriptive Statistics} \label{table:descriptives}
{\small \begin{tabular*}{\textwidth}{@{\extracolsep{\fill}}l@{}ccc}
    \toprule
    \textbf{Dataset} & \textbf{\makecell{Retailer A}} & \textbf{\makecell{Retailer B}} & \textbf{\makecell{Retailer C}} \\
    \midrule
    Total number of customers & 4,785 & 4,661 & 5,843 \\[5pt]
    Acquisition cohort definition\textsuperscript{a} & 12 months & 24 months & 24 months \\[5pt]
    Estimation period & 2 years & 2 years & 2 years \\[5pt]
    Holdout period& 4 years & 3 years & 4 years \\[5pt]
    Zero repeaters & 55.4\% & 48.1\% & 27.6\% \\[5pt]
    Number of purchases\textsuperscript{b}  & 19,854 & 19,938 & 20,628 \\[5pt]
    \makecell[l]{Average number of transactions\textsuperscript{b} \\ (standard deviation)} & \makecell{4.15\\(5.76)} & \makecell{4.28\\(4.60)} & \makecell{3.53\\(3.19)} \\[10pt]
    \makecell[l]{Average transaction value\textsuperscript{b} \\ (standard deviation)} & \makecell{75.99\\(138.20)} & \makecell{338.19\\(496.11)} & \makecell{40.14\\(66.23)} \\[10pt]
    \makecell[l]{Average interpurchase time\textsuperscript{b} \\  (standard deviation)} & \makecell{33.56 \\(22.13)} & \makecell{54.26\\ (50.49)}& \makecell{21.65\\ (16.21)} \\[10pt]
    Publicly available & yes  & yes  & yes  \\
    \bottomrule
\end{tabular*}}
\raggedright\scriptsize{~} \\
\raggedright\scriptsize{Note: \textsuperscript{a} The acquisition cohort definition indicates the time bin used to group customers by acquisition date (e.g., 12-month cohorts). \textsuperscript{b} Quantities measured for the estimation period.}
\end{table}%

Table \ref{table:descriptives} provides a summary of key descriptive statistics for all datasets. Importantly, the datasets span regimes with high one-time purchase rates, long interpurchase times, and high spend dispersion, directly reflecting the sparse and heterogeneous conditions under which long-horizon revenue forecasting is most difficult. For example, the inter-purchase time varies across the datasets between 21 and 54 days, while at the same time the average transaction value varies between 40 and 338 USD. Also, the descriptive statistics for zero-repeaters, i.e. customers who only purchase a single time, vary considerably between 27\% and 55\%. For all datasets, purchase records are available for multiple years, which enables testing of different prediction horizons. We conclude that this dataset selection facilitates robust benchmarking.
% It becomes clear that the diverse sample of purchase records from different industries indeed covers a wide range of transaction, attrition, and spending behavior.

%%%%%%%%%%%%%%%%%%%%%%%%%%%%%%%%%%%%%%%%%%%%%%%%%%%%%%%%%%%%%%%%%%%%%%

\subsection{Results}

We present the results of our benchmarking in two parts. First, we assess the predictive performance for our proposed approach for scenarios where covariate data is not available. Second, we leverage a case study that mimics a real-world customer base analysis application to assess the predictive performance of our proposed approach for scenarios where covariate data is available. This structure isolates the value of (i) flexible latent heterogeneity modeling in the absence of covariates and (ii) the incremental predictive value of incorporating covariates for a common use case.

The first benchmarking focuses on comparing four models over different prediction horizons. For all models, we compare the predictive performance for customer-level revenue prediction based on the root-mean-square error metric for each holdout period. Each model is trained on two years of data (see Table~\ref{table:descriptives}). As mentioned above, in addition to our novel approach, we also estimate the Pareto/NBD model combined with the GG model (PNBD + GG), the LSTM model combined with the GG model (LSTM + GG), and a joint model based on the work of \citet{GladyLemmensCroux2015} (PNBDGGII). None of these models uses customer-level covariates. However, the LSTM implementation automatically adds seasonality features, so it effectively benefits from additional inputs relative to the other benchmarks. Further, note that using models without any covariates is still a common choice due to privacy regulations, data quality issues (e.g., high number of missing values), or the requirement to build less complex predictive models in a shorter period of time.  

% Result Table - Model Without Covariates
\newlength{\numcol}
\setlength{\numcol}{0.12\linewidth}   % Adjust this width as needed
\begin{table}[ht!]
\centering
\renewcommand{\arraystretch}{1.2}
\caption{Benchmarking of models without covariates (RMSE customer-level revenue predictions)
\label{table:results-ind}
}
%\vspace{0.2cm}
{\small \begin  {tabular}{p{1.65cm}p{5.25cm}*{4}{>{\raggedleft\arraybackslash}p{\numcol}}}
       \toprule
       \multirow{2}{*}{\textbf{Dataset}} & \multirow{2}{*}{\textbf{Model}} & \multicolumn{4}{c}{\textbf{Prediction Periods (Weeks)}} \\
       & & 52 & 104 & 156 & 208\\
       \midrule
    \multirow{4}{*}{\shortstack[l]{Retailer\\ A}}
         & (1) PNBD + GG &  119.59 & 219.75 & 336.24 & 459.58\\
         & (2) LSTM + GG & 105.58~ & 173.67 & 262.14 & 359.41\\
         & (3) PNBDGGII  & 126.57~ & 188.90 & 276.22  & 369.22 \\
         & (4) CLVAE & \textbf{102.00}~ & \textbf{165.09} & \textbf{256.23} & \textbf{352.40}\\[4pt]
       \cdashline{1-6} \noalign{\vskip 4pt} % Dashed line only over number columns
    \multirow{4}{*}{\shortstack[l]{Retailer\\ B}}
         & (1) PNBD + GG  & 578.07 & 968.47 & 1,328.44& - \\
         & (2) LSTM + GG & 612.28 & 1,069.45 & 1,529.45 & -  \\
         & (3) PNBDGGII  & 628.68~ & 975.35 & 1,300.08 & -\\
         & (4) CLVAE & \textbf{565.90} & \textbf{945.12} & \textbf{1,291.00} & -\\[4pt]
    \cdashline{1-6} \noalign{\vskip 4pt} % Dashed line only over number columns
    \multirow{4}{*}{\shortstack[l]{Retailer\\ C}}
         & (1) PNBD + GG & 102.58 & 188.95 & 265.00 & 342.42 \\
         & (2) LSTM + GG & 103.01 & 190.48 & 259.32 & 321.99 \\
         & (3) PNBDGGII  & 113.01~ & 206.61 & 279.69 & 344.62\\
         & (4) CLVAE & \textbf{98.73} &\textbf{175.84} & \textbf{236.70} & \textbf{293.35}\\[2pt]
    \bottomrule
\end{tabular}
}
\raggedright\scriptsize{~} \\
\raggedright\scriptsize{Note: RMSE is computed across customers using cumulative realized revenue over the indicated prediction horizon. The best performing model per scenario is highlighted in bold.}

\end{table}

The results of this first benchmarking are presented in Table~\ref{table:results-ind} and are complemented by Online Appendix~1, which reports mean absolute error (MAE). Overall, the evidence supports the proposed CLVAE (Model 4) as the most accurate approach in this no-covariates setting. Across datasets and prediction horizons, CLVAE delivers the lowest out-of-sample RMSE.  

Relative to the traditional two-stage baseline PNBD + GG (Model 1), CLVAE yields systematic improvements that persist across different revenue scales and dispersion levels. For Retailer~A, the gains widen substantially over longer horizons (e.g., 352.40 vs.\ 459.58 at 208 weeks), suggesting that jointly capturing heterogeneity in attrition, transaction timing, and spending improves long-term revenue forecasts even when using parsimonious RFM-style inputs rather than full transaction histories (see Appendix~\ref{app:appendix_4}). The same pattern is observed for Retailer~B, where revenue levels and dispersion are markedly higher. CLVAE remains best at each reported horizon (e.g., 1,291.00 vs.\ 1,328.44 at 156 weeks), suggesting that the advantage generalizes beyond comparatively low-variance settings. For Retailer~C, differences are small at the shortest horizon, but CLVAE becomes clearly superior in the medium and long term (104 to 208 weeks). 

Importantly, these improvements are not explained simply by adopting any joint specification or by relying on sequence-based prediction. The joint model benchmark (Model 3) does not close the gap to CLVAE and is often comparable to, or worse than, the PNBD + GG baseline, implying that coupling processes through a more rigid joint formulation is insufficient. Meanwhile, the LSTM + GG benchmark (Model 2), which is explicitly designed to take advantage of sequential purchase histories, can be competitive in some cases, but it still does not match CLVAE’s accuracy across prediction horizons. Taken together, the pattern is consistent with CLVAE’s flexible latent-variable structure capturing persistent customer-level heterogeneity across attrition, transaction timing and spending, translating into systematically improved customer-level revenue predictions, especially as uncertainty compounds over longer forecast windows. Notably, this conclusion is conservative with respect to the LSTM + GG benchmark, which benefits from additional engineered seasonality inputs, yet does not dominate CLVAE at longer horizons.

A second benchmarking focuses on assessing the ability to account for contextual information through covariates. To this end, we leverage a case study that looks at one of the key modeling challenges in customer base analysis, i.e., how to best account for heterogeneity between customers from different acquisition cohorts. Industry practice follows recommendations from scholarly research, where instead of looking at the entire customer base, possible observed heterogeneity is accounted for by distinguishing more homogeneous cohorts of customers. These cohorts are formed according to the time period a customer started buying from the firm. There are multiple ways to address this from a modeling perspective: 

\begin{itemize}
    \item A naive approach would ignore this readily available information and estimate a single model across the entire customer base, which comes with minimal modeling complexity but also ignores a well-founded source of observed customer heterogeneity. 
    \item An alternative fits models such as the Pareto/NBD and GG separately for each of these cohorts. This reflects the assumption that customers that start in the same period (e.g. month) are relatively similar, and thus might be captured well by a parametric model. However, not only does this increase computational complexity and technical debt due to the estimation of multiple models, but this approach also may pose a challenge when estimating separate models for customers that have only been observed for a few days, weeks, or even months. In the latter case, the available training data is limited, which may lead to convergence issues during estimation or unreliable results. However, this approach has the advantage that all model parameters can vary by customer acquisition cohort, allowing for a lot of flexibility to adapt to particular purchase patterns for these customer groups.
    \item Another alternative is to explicitly account for the information on customers' acquisition cohort in the model. In other words, a single model that accounts for a customer's assignment to a cohort with covariates is used to assess simultaneously all customers in a firm's customer base.  Dummy variables are used to indicate in which period a customer started the relationship with the business. Depending on the modeling approach, this information is incorporated in different ways: 
    (a) For the traditional modeling approaches, such as the combination of the Pareto/NBD and the GG model, the rate parameters $\alpha$ and $\beta$ are cohort-specific, while all other parameters are shared. While this approach reduces modeling complexity and technical debt, its ability to account for the observed customer heterogeneity is limited. 
    (b) For the LSTM model, cohort membership is incorporated as a set of time-invariant customer-level covariates that are replicated across all time steps, transformed through a dense layer, and then concatenated with the time-varying seasonality and transaction inputs before being processed by the recurrent layer—both during training and autoregressive forecasting.
    (c) Including covariates on customers' acquisition cohort membership in the CLVAE model provides an alternative that combines all advantages. Being a single model, the modeling complexity is low, and thus, the practical applicability is straightforward, while the way that covariates are considered within the model provides similar levels of flexibility than running individual cohort-level models.
\end{itemize}

This case study therefore tests whether cohort information is best handled via either operationally costly cohort-specific model estimation or a single model that can flexibly absorb cohort effects. We assess these alternatives for Retailer C, the largest among the three datasets. For this dataset, we define monthly cohorts. As the training data builds on a 24-month definition for the focal acquisition cohort (see Table~\ref{table:descriptives}), we define 24 monthly cohort indicators to control for the homogeneity of customers acquired in the same month. The cohorts are the foundation to apply the modeling alternatives described above. Results are shown in Table~\ref{table:results-ind2}, with the benchmarks from Table~\ref{table:results-ind} added for convenience (Model 1 \& 2). Model 3 mirrors the common operational workflow of fitting separate likelihood-based models by acquisition cohort, which increases maintenance burden and can be unstable for newly acquired cohorts with very little training data.

% Result Table - Model With Covariates
\setlength{\numcol}{0.08\linewidth} 
\begin{table}[ht!]
\centering
\renewcommand{\arraystretch}{1.2}
\caption{Case study on models with covariates (RMSE customer-level revenue predictions; Retailer C)
\label{table:results-ind2}
}
%\vspace{0.2cm}
{\small \begin  {tabular}{p{9.8cm}*{4}{>{\raggedleft\arraybackslash}p{\numcol}}}
       \toprule
       \multirow{2}{*}{\textbf{Model}} &  \multicolumn{4}{c}{\textbf{Prediction Periods (Weeks)}} \\
        & 52 & 104 & 156 & 208\\
       \midrule
       (1) PNBD + GG    &102.58 & 188.95 & 265.00 & 342.42 \\
       (2) CLVAE  &98.73  &  175.84 & 236.70 & 293.35\\
       (3) PNBD + GG (separate model for each acquisition cohort) & 103.75 & 190.79 & 265.79 & 338.17\\
       (4) LSTM with covariates + GG & 102.36 & 184.00 & 248.87 & 307.31\\
       (5) PNBDGGII  with covariates  & 111.75~ & 204.40 & 276.46 & 340.12\\
       (6) PNBD with covariates + GG & 104.75 & 192.59 & 270.69 & 350.09\\
       (7) CLVAE with covariates & \textbf{98.32} &\textbf{173.74} & \textbf{233.50} & \textbf{284.11}\\[2pt]
\bottomrule
\end{tabular}}
\raggedright\scriptsize{~} \\
\raggedright\scriptsize{Note: RMSE is computed across customers using cumulative realized revenue over the indicated prediction horizon. The best performing model per scenario is highlighted in bold.} 
\end{table}
\vspace{-0.4cm}

Overall, the results show that cohort indicators can improve long-term revenue predictions, but the gains depend strongly on whether the model can translate cohort differences into flexible changes in customers' attrition, transaction, and spending processes. The CLVAE with covariates (Model~7) is the best-performing specification across all four horizons, and it improves consistently over the same architecture without covariates (Model~2). This pattern indicates that cohort membership adds an incremental predictive signal beyond what is captured by the baseline CLVAE specification. In contrast, fitting separate PNBD + GG models (Model 3) by monthly cohorts yields no gain over the single PNBD + GG model and is slightly worse across prediction horizons, indicating that cohort splitting alone does not translate into better long-horizon revenue forecasts in this setting. In this setting, CLVAE with covariates delivers the accuracy benefits one would seek from cohort-specific modeling while retaining a single-model deployment workflow.

The results also clarify why this advantage is specific to the proposed approach rather than a generic “covariates help” effect. Adding cohort indicators to traditional parametric specifications yields limited improvements and can even degrade performance when heterogeneity is forced through a narrow parameter subset (Model~6) or when estimation becomes more fragile in richer joint specifications (Model~5). The LSTM model with covariates (Model~4) improves over the covariate-free LSTM benchmark from Table~\ref{table:results-ind}, but it remains materially less accurate than CLVAE with covariates. This applies especially to longer horizons, suggesting that the proposed model extracts more of the cohort signal in the regime most relevant for long-run value assessment.

Taken together, the evidence suggests that acquisition-cohort membership contains a relevant predictive signal for long-term revenue. However, fully exploiting it requires a model that can flexibly propagate cohort effects through the full set of latent processes governing customers' attrition, transaction timing and spending. The covariate-augmented CLVAE provides this combination: a single deployable model with cohort-aware flexibility that delivers the most accurate predictions across all forecast horizons in this case study.

Summing up the two benchmark studies, CLVAE improves customer-level long-horizon revenue prediction across datasets and forecast horizons in non-contractual settings. This supports its value when transaction histories are sparse and irregular and heterogeneity modeling is consequential.

%%%%%%%%%%%%%%%%%%%%%%%%%%%%%%%%%%%%%%%%%%%%%%%%%%%%%%%%%%%%%%%%%%%%%%
%%%%%%%%%%%%%%%%%%%%%%%%%%%%%%%%%%%%%%%%%%%%%%%%%%%%%%%%%%%%%%%%%%%%%%
%%%%%%%%%%%%%%%%%%%%%%%%%%%%%%%%%%%%%%%%%%%%%%%%%%%%%%%%%%%%%%%%%%%%%%

\section{Discussion}\label{sec:Discussion}

This paper addresses a core problem in customer base analysis: forecasting long-horizon customer revenue from sparse and irregular transaction histories in non-contractual settings. We show that a VAE can bypass restrictive heterogeneity assumptions in traditional probabilistic models while preserving their tripartite process structure. CLVAE keeps the Empirical-Bayes anchoring that gives traditional models stability, but gains flexibility through amortized inference that learns a nonparametric heterogeneity distribution. As a result, this hybrid design provides robust, accurate, and scalable predictions, even with parsimonious data input, while also being able to account for a broad set of covariates if available. Given these properties, our novel approach is applicable to a diverse set of non-contractual business settings. Managerially, this supports a single-model deployment workflow for long-horizon revenue prediction, avoiding the operational burden of maintaining separate cohort-specific models while still exploiting cohort information if available.

Our empirical results provide evidence that the combination of the VAE framework with traditional probabilistic models advances customer base analysis. Previous literature has identified several advantageous features for any customer base model that have not yet been available in a single approach. Leveraging techniques from traditional probabilistic modeling and deep learning, our approach brings together isolated streams within the literature on customer base analysis. Thereby, the proposed approach does not just simply use the standard VAE framework but carefully tailors it to traditional probabilistic customer base models that have stood the test of time, namely the Pareto/NBD model and the GG model. As a consequence, the approach yields an approximate posterior over customer-level latent propensities that represent customers' attrition, transaction, and spending processes. Thereby, we extend strictly parametric mixing assumptions by learning a flexible latent representation of heterogeneity. This design balances process discipline with flexible heterogeneity modeling, improving stability in sparse regimes while avoiding purely autoregressive overfitting. In addition, the proposed approach allows modeling the full dependence structure between the individual processes. 

In summary, CLVAE improves out-of-sample customer-level revenue forecasts relative to traditional probabilistic approaches and competitive deep-learning across datasets and prediction horizons. Critically, it does so with a single deployable model that remains data-parsimonious when only transaction histories are available and can incorporate covariates when firms have access to such information and are permitted to use it. This combination of process-based structure with a learned heterogeneity distribution explains why gains persist as the forecast horizon lengthens.

Compared to the application of an LSTM model as in \cite{ValendinReuttererPlatzerKalcher2022}, our approach allows retaining some of the structure of the widely successful Pareto/NBD model, while also allowing for a natural inclusion of the spending process via the GG model. In other words, it still builds on the same economic primitives that defined the “gold standard" in customer base analysis for several decades \citep{Jerath2011} and thus contrasts itself to standard autoregressive forecasting approaches. The natural split into customers' attrition, transaction and spending process is still present in the CLVAE model. This not only makes our model more data parsimonious, working well with only the RFM data (in contrast to the full transaction history), but can also lead to improved forecasts, as shown in Section \ref{sec:EmpiricalApplication}. On a technical level, we note that, while the approach of \cite{ValendinReuttererPlatzerKalcher2022} necessitates discretization of time, the CLVAE model builds on a process that is still truly continuous with purchases potentially happening at any time point.

Several avenues emerge for future research. First, while we demonstrate performance across multiple real-world datasets, an extension would be to include time-varying covariates such as seasonality indicators. This would require defining time-varying latent variables in the model specification. However, the proposed model in its current form already outperforms deep-learning benchmarks that include such information (see Section~\ref{sec:EmpiricalApplication}). If the goal is predictive accuracy at the customer level rather than evaluating the customer cohort as a whole, the impact of adding such seasonality indicators, which vary only by calendar time but not by customer, may be limited \citep{BachmannMeiererNaef2021}. Second, further extensions have been proposed in customer base modeling that might be worth exploring in a VAE framework, e.g., regularity of purchase behavior \citep{PlatzerReutterer2016} or intra-customer process correlation \citep{GladyLemmensCroux2015}. While these extensions may be possible to accommodate in the VAE framework, they also come with more data requirements. Beyond data on how recently customers purchased, how often customers bought, and how much they spent, they also require access to the complete transaction history. In conclusion, these directions for further research would strongly build, but not change, the fundamental contribution that we make with the present work. They may nevertheless provide further guidance for customer base analysis in specific business settings. 

In summary, our novel approach extends the toolkit of marketing researchers with a scalable likelihood-based modeling technique that learns and leverages latent customer heterogeneity for long-horizon revenue prediction in non-contractual settings. More broadly, we outline a way to bring VAE-style representation learning to research settings where process structure matters. Thus, the econometrically meaningful process structure can be retained rather than replaced them with a black-box mapping. To facilitate replication and further research, we released a Python implementation of our novel model along with code to reproduce our results in a public repository  (\emph{link removed to ensure anonymous review}).

%%%%%%%%%%%%%%%%%%%%%%%%%%%%%%%%%%%%%%%%%%%%%%%%%%%%%%%%%%%%%%%%%%%%%%
%%%%%%%%%%%%%%%%%%%%%%%%%%%%%%%%%%%%%%%%%%%%%%%%%%%%%%%%%%%%%%%%%%%%%%
%%%%%%%%%%%%%%%%%%%%%%%%%%%%%%%%%%%%%%%%%%%%%%%%%%%%%%%%%%%%%%%%%%%%%%

\newpage

%\THEEndNotes
%\begingroup \parindent 0pt \parskip 0.0ex \def\enotesize{\normalsize} \theendnotes \endgroup

% Appendix here
% Options are (1) APPENDIX (with or without general title) or
%             (2) APPENDICES (if it has more than one unrelated sections)
% Outcomment the appropriate case if necessary
%
% \begin{APPENDIX}{<Title of the Appendix>}
% \end{APPENDIX}
%
%   or
%
\appendix

\section{Neural network architecture}\label{app:appendix_1}

The proposed approach is a deep generative framework for modeling customer revenue, in particular, for deriving long-term prediction. Hereby, we generalize the combination of the Pareto/NBD and the GG model into a single architecture that captures nonlinearities and interactions in the latent posterior distribution. Key characteristics of the model have been discussed, in  Section~\ref{sub:model_spec}. For key details of the model architecture, see Figure~3.

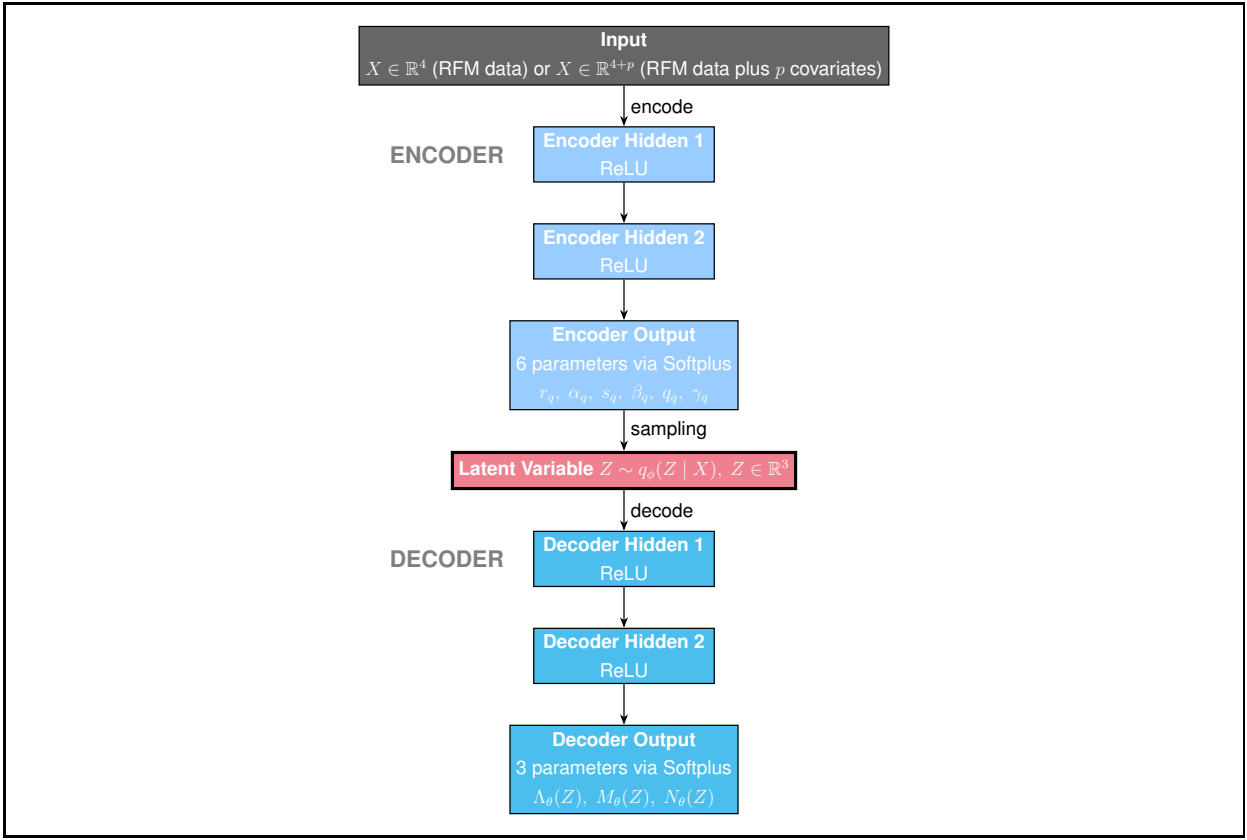
\begin{figure}[H]
\label{fig:vae_architecture}
\centering
\begin{tcolorbox}[colback=white, colframe=black, 
                  sharp corners, boxrule=1pt, left=10pt, right=10pt, top=5pt, bottom=5pt]
                  
\centering                  
\scalebox{0.60}{
\begin{tikzpicture}[
    node distance=0.9cm,
    every node/.style={font=\sffamily}, 
    layer/.style={rectangle, sharp corners, draw=black, thick, 
                  minimum height=0.8cm, minimum width=4.0cm, 
                  align=center, fill=#1, text=white},
    arrow/.style={->, >=Stealth, thick},
]

% ======================= 
% ENCODER 
% ======================= 
\node[layer=gray!120] (input) {\textbf{Input} \\ $X \in \mathbb{R}^{4}$ (RFM data) or $X \in \mathbb{R}^{4 + p}$ (RFM data plus $p$ covariates)
};

\node[layer=blue!100!black!50!cyan!40, below=of input] (enc1) 
    {\textbf{Encoder Hidden 1} \\ ReLU};

\node[layer=blue!100!black!50!cyan!40, below=of enc1] (enc2) 
    {\textbf{Encoder Hidden 2} \\ ReLU};

\node[layer=blue!100!black!50!cyan!40, below=of enc2] (encout) 
    {\textbf{Encoder Output} \\ 6 parameters via Softplus \\ 
     $r_q,\ \alpha_q,\ s_q,\ \beta_q,\ q_q,\ \gamma_q$};

% Arrow flow encoder
\draw[arrow] (input) -- node[right, font=\sffamily]{encode} (enc1);
\draw[arrow] (enc1) -- (enc2);
\draw[arrow] (enc2) -- (encout);

% ======================= 
% LATENT SPACE 
% ======================= 
\node[layer=red!50!purple!50, below=0.9cm of encout, line width=2pt] (latent) 
   {\textbf{Latent Variable} $Z \sim q_\phi(Z \mid X), \ Z \in \mathbb{R}^{3}$};

\draw[arrow] (encout) -- node[right, font=\sffamily]{sampling} (latent);

% ======================= 
% DECODER 
% ======================= 
\node[layer=cyan!90!blue!60, below=0.9cm of latent] (dec1) 
    {\textbf{Decoder Hidden 1} \\ ReLU};

\node[layer=cyan!90!blue!60, below=of dec1] (dec2) 
    {\textbf{Decoder Hidden 2} \\ ReLU};

\node[layer=cyan!90!blue!60, below=of dec2] (decout) 
    {\textbf{Decoder Output} \\ 3 parameters via Softplus \\ 
     $\Lambda_\theta(Z),\ M_\theta(Z),\ N_\theta(Z)$};

\draw[arrow] (latent) -- node[right, font=\sffamily]{decode} (dec1);
\draw[arrow] (dec1) -- (dec2);
\draw[arrow] (dec2) -- (decout);

% ======================= 
% ENCODER/DECODER LABELS 
% ======================= 
\node[left=0.5cm of enc1, font=\large\bfseries\sffamily, text=gray] {ENCODER};
\node[left=0.5cm of dec1, font=\large\bfseries\sffamily, text=gray] {DECODER};
\end{tikzpicture}
}

\end{tcolorbox}
\caption{Architecture of the CLVAE model}
\end{figure}

The architecture comprises an encoder and a decoder, both implemented as fully connected feedforward neural networks with ReLU activations. The encoder maps observed customer-level features into the parameters of latent Gamma distributions, thereby defining customer-level latent variables that capture heterogeneity in customer behavior. The decoder then transforms sampled latent variables into the parameters of the conditional likelihood over observed transactions, enabling probabilistic reconstruction of customer behavior .

The encoder maps the input $X$, which consists of RFM data and, if available, customer-level covariates, through two ReLU layers to produce six positive parameters $(r_q, \alpha_q, s_q, \beta_q, q_q, \gamma_q)$ via a Softplus output layer, with biases initialized from Gamma prior means. The encoder is parameterized by $\phi$, defining the variational posterior $q_\phi(Z|X)$. 

Next, latent variables $Z \in \mathbb{R}^{3}$ are sampled from this posterior and passed through the decoder, which consists of two ReLU layers and outputs three positive parameters $(\Lambda, M, N)$ via Softplus, parameterized by $\theta$, defining the conditional likelihood over observed customer transactions.

%{The encoder network consists of two fully connected layers with ReLU activations mapping the input features to the Gamma distribution parameters of each latent variable component. A deterministic auxiliary network $\Phi_\theta$ then transforms these samples into auxiliary latent variables~\eqref{eq:neural_net}. here the reference to the equation is broken. please fix and delete this comment Model training optimizes the evidence lower bound (ELBO), balancing reconstruction accuracy and latent space regularization. The dataset is randomly shuffled and divided into mini-batches at each epoch for computational efficiency. For each mini-batch, variational parameters are computed by the encoder, latent samples are drawn using the reparameterization trick, auxiliary latents are deterministically computed. The ELBO is computed as~\eqref{eq: elbo_pnbd_vae}. Gradients of the negative ELBO are backpropagated through all network components, and parameters $\phi$ and $\theta$ are updated via an adaptive optimizer such as Adam. Early stopping with a patience criterion based on validation ELBO is employed to prevent overfitting.

%\textcolor{blue}{Continue here by shortly describing the ELBO loss, the estimation procedure and add the algorithm environment, without too much details}

\section{Neural network training}\label{app:appendix_2}

The model parameters $(\phi, \theta)$ are trained by maximizing the Evidence Lower Bound (ELBO) via stochastic gradient descent. We adopt the canonical fitted parameters for the Pareto/NBD and the GG model as prior parameters. In this context, we note that the likelihood \eqref{eq:likelihood} also includes the parameter $p$. This parameter is fixed to the parameter obtained through the GG model. However, it is also possible to adapt a fully Bayesian approach where $p$ itself is endowed with a prior distribution, with the posterior approximated by the VAE. Given this setup, we encode the parameters of the variational latent Gamma distributions. Then, the samples are drawn using differentiable, reparameterizable Gamma sampling, leveraging stochastic primitives available in frameworks such as PyTorch~\citep{figurnov2018implicit}.

For computational scalability, the dataset is randomly shuffled and partitioned into mini-batches at each epoch. Since training is unsupervised, overfitting is monitored via the validation ELBO, and early stopping is applied with a predefined patience criterion. Architectural hyperparameters, including the number of hidden layers, units per layer, and latent dimensionality, must be tuned according to dataset size and complexity, balancing model expressiveness with generalization performance. To further improve stability in our model, we initialize the output-layer biases of the encoder using Gamma prior means and apply Softplus activations to ensure all inferred latent parameters remain positive. 

The model hyperparameters and training configuration used to produce the results reported in Section~\ref{sec:EmpiricalApplication} are summarized in Table~4.

\begin{table}[H]
\centering
\renewcommand{\arraystretch}{1.2}
\caption{Model hyperparameters and training configuration 
\label{fig:training-config}
}
\vspace{0.2cm}
{\small
\begin{tabular}{p{0.31\linewidth} p{0.32\linewidth} p{0.31\linewidth}}
\toprule
\textbf{Category} & \textbf{Item} & \textbf{Value} \\
\midrule
\multirow{5}{*}{\textit{\textbf{Model Hyperparameters}}}
 & Encoder hidden layers & 2 \\
 & Encoder units per layer & 64, 32 \\
 & Latent dimension & 3 \\
 & Decoder hidden layers & 2 \\
 & Decoder units per layer & 32, 64 \\[4pt]
\cdashline{1-3}\noalign{\vskip 4pt}
\multirow{8}{*}{\textit{\textbf{Training Configuration}}}
 & Optimizer & Adam \\
 & Learning rate & 0.001 \\
 & Loss & ELBO \\
 & Batch size & 64 \\
 & Max epochs & 1000 \\
 & MC samples & 10 \\
 & Early stopping patience & 100 \\
 & Seed & 50 \\
\bottomrule
\end{tabular}
}
\end{table}

\vspace{-0.3cm}

\section{Prediction routine}\label{app:appendix_3}

In the following, we suppress the index $i$, writing out the simulation for a ``generic'' customer and use the shorthand $\Lambda_\theta=\Lambda_\theta(\lambda, \mu, \nu)$, $M_\theta=M_\theta(\lambda, \mu, \nu)$ and $N_\theta=N_\theta(\lambda, \mu, \nu)$. We then repeatedly exploit that, conditional on $\Lambda_\theta, M_\theta, N_\theta$, the assumed data-generating process follows the distribution of the Pareto/NBD and GG model. Note that spending is independent of the number of transactions conditioned on $\Lambda_\theta, M_\theta, N_\theta$. Thus, if $S(t)$ is the total spending in the period $[T,T+t)$ and $N(t)$ the total number of transactions in $[T,T+t)$, adapting the derivations in \cite{Fader2005a, Fader2013a}, we seek to approximate: 
\begin{align*}
    &E[S(t) \mid r, \alpha, s, \beta, x, t_x, T] \\
    &= \int_0^\infty \int_0^\infty \int_{T}^{\infty} E[\sum_{j=1}^{N(t)} S_j \mid N_{\theta},\Lambda_{\theta}, \omega] \\
    &f(\omega \mid M_{\theta}, \Omega > T) d\omega P(\Omega > T \mid \Lambda_{\theta}, M_{\theta}, x, t_x, T) \\
    &g(\Lambda_\theta, M_\theta, N_{\theta} \mid x, t_x, T, \bar{z})  d\Lambda_\theta dM_\theta d N_{\theta}.
\end{align*}
In the above, $E[\sum_{j=1}^{N(t)} S_j \mid N_{\theta},\Lambda_{\theta}, \omega]$ is the expectation over a random sum with $N(t)$ elements, whereby $N(t)$ is the number of events in a Poisson process with parameter $\Lambda_{\theta}$ on $[T,\min(T+t, \omega)]$, and $S_j \sim \text{Gam}( q, N_{\theta})$, and
\begin{align}\label{paliveappendix}
    &P(\Omega > T \mid \Lambda_{\theta}, M_{\theta}, x, t_x, T) \nonumber \\
    &=\left(1 +\frac{M_{\theta}}{\Lambda_{\theta} + M_\theta} \left( \exp\big[(\Lambda_{\theta} + M_{\theta})(T - t_{x})\big] - 1 \right) \right)^{-1},
\end{align}
adapted from \cite[Section 5]{Fader2005a}. By the memorylessness property of the exponential distribution, it follows that $f(\omega \mid M_{\theta}, \Omega > T)$ is the density of an exponential distribution itself with parameter $M_{\theta}$.

We simulate expected number of purchases and expected total spending, for all customers using fully vectorized Monte Carlo simulations over latent variables $Z = (\lambda, \mu, \nu)$, and horizons $t_1, \ldots, t_K$. Let $t_{max}=\max(t_1, \ldots, t_K)$. Though, we consider a generic customer here, whenever possible calculations are vectorized over customers, simulations, and horizons.

  \begin{enumerate}[label=\arabic*., listparindent=1.5em]
   \item{\textbf{Sample latent variables:} For $l = 1,\ldots,L$ Monte Carlo samples, draw
    \[
    \lambda^{l} \sim \text{Gam}(\hat r, \hat \alpha), 
    \mu^{l} \sim \text{Gam}(\hat s, \hat \beta), 
    \nu^{l} \sim \text{Gam}(\hat q, \hat \gamma),
    \]
    where the hyperparameters 
    \[
    (\hat r, \hat \alpha, \hat s, \hat \beta, \hat q, \hat \gamma, \hat p) = f_\phi([x,t_x,T , \mathbf{Y}])
    \] 
    are outputs of the proposed neural network $f_\phi$, optimized on the dataset, with $\mathbf{Y}$ representing covariates. }

    \item \textbf{Combine latent posterior using the decoder:} For each sample $l$, compute
    \[
    (\Lambda_\theta^l, M_\theta^l, N_\theta^l) = \Phi_\theta(\lambda^l, \mu^l, \nu^l),
    \]
where $\Phi_\theta$ is the encoder neural network optimized on the dataset.
    \item \textbf{Compute survival probability:}
    Draw a Bernoulli mask for each customer and using \eqref{paliveappendix}, simulating the event $\mathbf{1}\{\Omega > T\}$ given $\Lambda_\theta^{l}, M_\theta^{l}, x, t_{x}, T$.
    % \[
    % p_{\text{alive}} = \frac{1}{1 + \frac{M_\theta}{\Lambda_\theta + M_\theta} \left( \exp[(\Lambda_\theta + M_\theta)(T - t_{x})] - 1 \right)}, 
    % \quad p_{\text{alive}} \in [0,1].
    % \]

    %\item \textbf{Sample alive status:} 

    \item \textbf{Sample dropout times:}  If the customer has $\mathbf{1}\{\Omega > T\}=1$, draw
    \[
    \Omega^l \sim T + \text{Exp}(M_\theta),
    \]
otherwise if $\mathbf{1}\{\Omega > T\}=0$ (customer is inactive after $T$), set $\Omega^l=T$.
    \item \textbf{Simulate cumulative interpurchase times:}  Generate interpurchase times:  
        \[
        t_{x+j-1, x+j}^l \sim \text{Exp}(\Lambda_\theta), \quad j = 1,\ldots,N_\text{max},
        \]
        and compute transaction times:
        \[
        t_{x,x+j}^{l} = T + \sum_{\ell=1}^{j} t_{x+\ell-1, x+\ell}^l,
        \]
        as long as $t_{x,x+j}^{l}\leq \min(T+t_{max}, \Omega^l).$

    \item \textbf{Multiple horizons:} For a list of forecast horizons $t_1, \ldots, t_K$, count valid purchases per horizon by checking which cumulative times fall below each horizon and are before dropout. This leads to a number $N^{(l)}(t_k)$ of simulated transactions for simulation $l$ and horizon $t_k$.

    % \item \textbf{Simulate spending per purchase:}  
    % For customer $i$, sample spending from a Gamma distribution:
    % \[
    % S_i^{(s)} \sim \text{Gam}(\text{shape}=p \cdot N_i, \text{rate}=q \cdot N_i),
    % \]
    % using vectorized operations over all customers, simulations, and horizons.

    \item \textbf{Simulate total spending:}  
    Sample spending from a Gamma distribution for simulation $l$ and horizon $t_k$:
    \[
    S^{(l)}(t_k) \sim \text{Gam}(\text{shape}=p \cdot N^{(l)}(t_k), \text{rate}=q).
    \]
    %using vectorized operations over all customers, simulations, and horizons.

    \item \textbf{Compute expected values:}  
    Average over $L$ simulations to get expected purchases, spending for each customer and horizon: \newline
%\begin{align*}
%    \text{E}[N(t_k)] &= \frac{1}{L} \sum_{l=1}^L N^{(l)}(t_k),\\
%    \text{E}[S(t_k)] &= \frac{1}{L} \sum_{l=1}^L S^{(l)}(t_k). 
%\end{align*}
\noindent\resizebox{\linewidth}{!}{
\begin{minipage}{\linewidth}
\begin{align*}
    \text{E}[N(t_k)] &= \frac{1}{L} \sum_{l=1}^L N^{(l)}(t_k),\\
    \text{E}[S(t_k)] &= \frac{1}{L} \sum_{l=1}^L S^{(l)}(t_k). 
\end{align*}
\end{minipage}}

    \item \textbf{Vectorization and scalability:} All steps (survival, dropout, interpurchase, spending) are implemented using fully vectorized tensor operations, enabling efficient simulation for large numbers of customers and Monte Carlo samples without loops.
\end{enumerate}

We note that certain steps in this simulation could be simplified with closed-form expressions. However, speed gains appear to be modest and the previous formulation leaves room to allow for more general models.

\section{Model extensions: Leveraging additional information in customers' transaction data}\label{app:appendix_4}

% Invisible notes, do not delete: We use the following structure for the encoder right now:
% %%%%
% Purchase History → LSTM → FC1 → FC2 → Output → 6 Parameters
%                    (2→4) (4→64) (64→32) (32→6)
% %%%%
% This means the LSTM really is part of the encoder together with two more layers! However, we might also see this as first compressing the transaction data into 4 dimensions, just as we do manually!

For a generic customer, let  $t_{0}, t_{1}, \ldots, t_{x}$ be the transaction times with associated spending values $z_0, \ldots, z_x$. Here, $(t_0, z_0)$ correspond to the timing and the spending of the first transaction (i.e., a customer becoming "alive").

As discussed in the main text, the CLVAE outlined in Section \ref{sub:model_spec}, only considers RFM indicators $(x,t_{x}, T, \bar{z})$. This is natural for the probabilistic decoder, as the likelihood of the Pareto/NBD model combined with the likelihood of the GG model is a function of these data only, as seen in Equation \eqref{eq:likelihood}. Moreover, it can be shown that the posterior distribution of $\lambda, \mu, \nu$ conditioned on the transaction history also depends on $(x,t_{x}, T, \bar{z})$ only. Since the encoder approximates this posterior distribution, this motivated our approach in the main text. However, in the VAE framework the probabilistic encoder might benefit from having access to the full transaction history to better approximate the posterior distribution. 

As such, we build here a model extension, whereby we let the decoder depend on the interpurchase times $\mathbf{t}=(t_{0}, t_{0,1}, \ldots, t_{x-1,x})$, with $t_{j-1,j}=t_{j}-t_{j-1}$ and spending $\mathbf{z}=(z_{0}, z_{1}, \ldots, z_{x})$. The encoder now uses $X=[\mathbf{t}, \mathbf{z}]$,
% \begin{align*}
% X=[\mathbf{t}, \mathbf{z}]=\begin{pmatrix}
%     t_{0} & z_0\\
%     t_{0,1} & z_1\\
%     \vdots&\vdots\\
%     t_{x-1,x} & z_x
% \end{pmatrix},    
% \end{align*}
 for a given customer to create the parameters $\alpha(X),r(X),\beta(X), s(X), \gamma(X), q(X)$. This is used to sample the posterior $(\lambda, \mu, \nu)$ and then fed into the same decoder as before. In practice, $X_i$ now has different lengths for each customer $i$. To deal with this, we use a bivariate LSTM model in the first layers of the encoder. We refer to the model in the main text as CLVAE (RFM) and to the model using the LSTM as CLVAE (LSTM). Importantly, we also considered the somewhat more complex Transformer architecture \citep{Vaswani2017} in place of the LSTM architecture, but did not find consistent improvements across our datasets that would justify this computationally more intensive approach.

%, as illustrated in Figure \ref{Modelview_2}. 

%Finally, we note that how to add covariates in this case needs to be assessed carefully and we leave this for further work. 

% FIGURE 2
% \begin{figure}[h!]
% \begin{center}
% \includegraphics[clip, trim=1.2cm 0.4cm 1.2cm 0.4cm, width=16.5cm]{figures/vae-model.pdf}
% \caption{Conceptual Visualization of the Basic Model Structure using the full transaction history. \textcolor{red}{This figure is a placeholder and needs to be updated}}
% \label{Modelview_2}
% \end{center}
% \end{figure}

% Result Table - Model Without Covariates
%\newlength{\numcol}
\setlength{\numcol}{0.12\linewidth}   % Adjust this width as needed
\begin{table}[b!]
\centering
\renewcommand{\arraystretch}{1.2}
\caption{Benchmarking of additional models without covariates (RMSE customer-level revenue predictions)
\label{table:results-ind_appendix}
}
%\vspace{0.2cm}
{\small \begin  {tabular}{p{1.55cm}p{5.3cm}*{4}{>{\raggedleft\arraybackslash}p{\numcol}}}
       \toprule
       \multirow{2}{*}{\textbf{Dataset}} & \multirow{2}{*}{\textbf{Model}} & \multicolumn{4}{c}{\textbf{Prediction Periods (Weeks)}} \\
       & & 52 & 104 & 156 & 208\\
       \midrule
    \multirow{2}{*}{\shortstack[l]{Retailer\\ A}}
         & (2) CLVAE (RFM) & 102.00~ & 165.09 & 256.23 & 352.40\\
        & (5) CLVAE (LSTM) & \textbf{99.82}~ & \textbf{161.15} & \textbf{249.24} & \textbf{338.86}\\[4pt]
       \cdashline{1-6} \noalign{\vskip 4pt} % Dashed line only over number columns
    \multirow{2}{*}{\shortstack[l]{Retailer\\ B}}
         & (4) CLVAE (RFM)& \textbf{565.90} & \textbf{945.12} & \textbf{1291.00} & -\\
        & (5) CLVAE (LSTM) & 567.53~ & 948.78 & 1299.21 & -\\[4pt]
    \cdashline{1-6} \noalign{\vskip 4pt} % Dashed line only over number columns
    \multirow{2}{*}{\shortstack[l]{Retailer\\ C}}
         & (4) CLVAE (RFM) & \textbf{98.73} &\textbf{175.84} & \textbf{236.70} & \textbf{293.35}\\
        & (5) CLVAE (LSTM) & 104.29~ & 190.92 & 260.53 & 322.71\\[2pt]
    \bottomrule
\end{tabular}
}
\raggedright\scriptsize{~} \\
\raggedright\scriptsize{Note: The best performing model per scenario is highlighted in bold.}

\end{table}

There are two essential ways to look at CLVAE (LSTM) in comparison to CLVAE (RFM): First, the encoder is now simply a neural net capable of taking the full transaction history as an input. Second, in our CLVAE (LSTM) architecture, we concatenate an LSTM model with the layers used in the encoder of the original CLVAE (RFM). As such, while we manually compress the full transaction data into the RFM data $(x,t_{x}, T, \bar{z})$ for the CLVAE~(RFM), the CLVAE~(LSTM) learns the compression in the first step through an LSTM component. In essence, this corresponds to exchanging a manual compression for one that is learned by the model, bringing advantages and disadvantages simultaneously. 

These considerations are reflected in the results of an additional benchmarking shown in  Table \ref{table:results-ind_appendix}. While we are better than the original model for Retailer~C, we are somewhat worse in the other two, at least with our chosen tuning parameters. A crucial difficulty in the full model lies in the fact that more tuning parameters must be chosen, in addition to a higher estimation error.

\newpage

%%%%%%%%%%%%%%%%%%%%%%%%%%%%%%%%%%%%%%%%%%%%%%%%%%%%%%%%%%%%%%%%%%%%%%
%%%%%%%%%%%%%%%%%%%%%%%%%%%%%%%%%%%%%%%%%%%%%%%%%%%%%%%%%%%%%%%%%%%%%%
%%%%%%%%%%%%%%%%%%%%%%%%%%%%%%%%%%%%%%%%%%%%%%%%%%%%%%%%%%%%%%%%%%%%%%

% Acknowledgments here
% \ACKNOWLEDGMENT{We would like to express our sincere gratitude to \textit{|anonymized for review|} for their invaluable contributions to this research. We are also grateful to \textit{|anonymized for review|} for their support and assistance throughout the course of this work.}
% \newpage

% References here (outcomment the appropriate case)

% CASE 1: BiBTeX used to constantly update the references
%   (while the paper is being written).
%\bibliographystyle{informs2014} % outcomment this and next line in Case 1
%\bibliography{<your bib file(s)>} % if more than one, comma separated

%\bibliographystyle{informs2014} % outcomment this and next line in Case 1
%\bibliography{sample} % if more than one, comma separated

% CASE 2: BiBTeX used to generate mypaper.bbl (to be further fine tuned)
%\input{mypaper.bbl} % outcomment this line in Case 2

%If you don't use BiBTex, you can manually itemize references as shown below.

%\bibliographystyle{nonumber}

%%%%%%%%%%%%%%%%%%%%%%%%%%%%%%%%%%%%%%%%%%%%%%%%%%%%%%%%%%%%%%%%%%%%%%
%%%%%%%%%%%%%%%%%%%%%%%%%%%%%%%%%%%%%%%%%%%%%%%%%%%%%%%%%%%%%%%%%%%%%%
%%%%%%%%%%%%%%%%%%%%%%%%%%%%%%%%%%%%%%%%%%%%%%%%%%%%%%%%%%%%%%%%%%%%%%

\bibliographystyle{informs2014}
\bibliography{references.bib}

\end{document}